\pdfoutput=1

\documentclass[11pt]{article}

\usepackage[preprint]{acl}

\usepackage{times}
\usepackage{latexsym}
\usepackage{amsfonts}
\usepackage[T1]{fontenc}

\usepackage[utf8]{inputenc}

\usepackage{microtype}

\usepackage{inconsolata}

\usepackage{graphicx}
\usepackage{dashrule} 
\usepackage{subcaption}

\usepackage{tcolorbox}
\usepackage{duckuments}
\usepackage{adjustbox}
\usepackage{booktabs}
\usepackage{amsmath} 
\usepackage{enumitem}
\usepackage{cleveref}
\usepackage{multirow} 

\title{Lost in the Prompt Order: \\ Revealing the Limitations of Causal Attention in Language Models}

\author{
Hyunjong Ok$^{1,2}$ \quad Jaeho Lee$^1$ \\
$^1$POSTECH \qquad $^2$HJ AILAB \\ 
\texttt{hyunjong.ok@gmail.com, jaeho.lee@postech.ac.kr}
}

\begin{document}
\maketitle
\begin{abstract}
Large language models exhibit surprising sensitivity to the structure of the prompt, but the mechanisms underlying this sensitivity remain poorly understood. In this work, we conduct an in-depth investigation on a striking case: in multiple-choice question answering, placing context before the questions and options (CQO) outperforms the reverse order (QOC) by over 14\%p, consistently over a wide range of models and datasets. Through systematic architectural analysis, we identify causal attention as the core mechanism: in QOC prompts, the causal mask prevents option tokens from attending to context, creating an information bottleneck where context becomes invisible to options. 
\end{abstract}

\section{Introduction}
Large language models (LLMs) are highly sensitive to prompt structure. Even minor changes in surface forms---such as instruction phrasing, example placements, or reasoning elicitation---can lead to substantial differences in the prediction quality of the models \cite{wei2022chain, kojima2022large}.

Despite the importance of this sensitivity for the practical reliability of LLMs, our current understanding remains largely descriptive. We know ``what'' LLMs are sensitive to, but not ``why.''
For instance, \citet{lu2022fantastically} has shown that permuting the order of demonstrations in in-context learning (ICL) can dramatically affect accuracy, yet offers limited insight into what makes certain orders preferable.
Likewise, recent studies on multiple-choice question answering (MCQA) report significant performance fluctuations under option permutations, but stop short of identifying the mechanisms underlying this phenomenon \citep{pezeshkpour-hruschka-2024-large, zheng2024large}. 

\begin{figure}[t]
    \centering
    \includegraphics[width=1\linewidth]{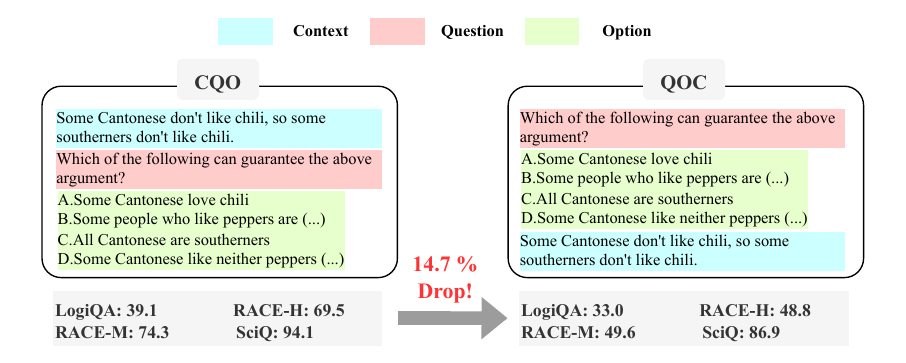}
    \caption{\textbf{Performance gap between CQO and QOC.} We measure the average accuracies of 21 decoder-only LLMs on 4 different datasets, when prompted in two distinct structures; CQO (context-question-option) and QOC (question-option-context).}
    \label{fig:fig2}
\end{figure}

In this work, we undertake an in-depth study of a less obvious but consequential form of prompt-sensitivity: the ordering of components in MCQA prompts.
A typical MCQA prompt consists of three elements: a context passage (C), a question (Q), and a set of options (O).
Intuitively, reordering these components should have little effect, as their semantic content remains unchanged. Contrary to this expectation, we find that placing the context before the questions and options (CQO) consistently and substantially outperforms the reverse ordering (QOC) across a wide range of setups (\Cref{fig:fig2}).\footnote{A similar phenomenon has also been noted by \citet{shaier2024not} outside the MCQA context, which also does not demystify why such phenomenon happens.} 

\begin{figure}[t]
\centering

\captionsetup[subfigure]{justification=centering}

\begin{subfigure}[t]{\linewidth}
  \centering
  \includegraphics[width=0.95\linewidth,trim=0 0 0 0,clip]{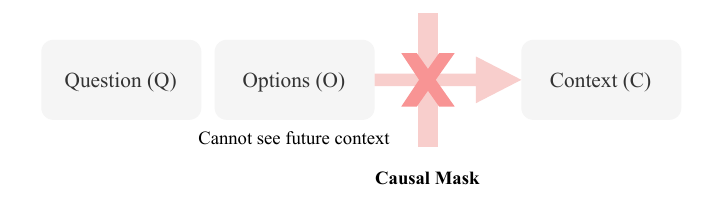}
  \subcaption{Decoder model}\label{fig:fig1_a}
\end{subfigure}

\begin{subfigure}[t]{\linewidth}
  \centering
  \includegraphics[width=0.95\linewidth,trim=0 0 0 0,clip]{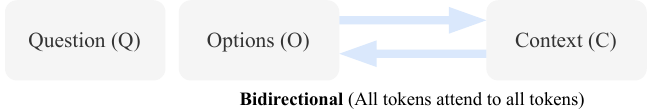}
  \subcaption{Encoder model}\label{fig:fig1_b}
\end{subfigure}

\caption{\textbf{Decoder vs. Encoder attention.}
In QOC (Question$\rightarrow$Options$\rightarrow$Context), causal masking prevents decoder models from attending to the context while selecting among options, so they often answer from option priors rather than evidence. Encoder models use bidirectional attention and can condition on the context when scoring the options.}
\label{fig:fig1}
\end{figure}

To explain this phenomenon, we formulate three competing hypotheses and evaluate them through a series of carefully controlled experiments. By systematically validating (or invalidating) each hypothesis, we aim to uncover the underlying factors that drive context-order sensitivity in MCQA. Specifically, we consider the following hypotheses:

\begin{itemize}[leftmargin=*,topsep=0pt,parsep=0pt]
\item \textbf{Hypothesis 1: Biased training data.} CQO-style prompts may be more prevalent in training data, making QOC an unfamiliar format for the model.
\item \textbf{Hypothesis 2: Failures in option recall.} QOC structure makes it difficult for the model to recall the options located in the middle of the prompt, a.k.a. ``lost-in-the-middle'' \citep{liu2024lost}.
\item \textbf{Hypothesis 3: Causal attention (winner).} The causal attention structure in decoder-only transformers makes it impossible for the option tokens to attend to the context tokens, creating an information bottleneck (\Cref{fig:fig1}).
\end{itemize}




Our experiments---conducted on a wide range of datasets and models --- support the third hypothesis, and rule out the other two hypotheses. Stepping further, we identify several factors that can affect the impact of causal masking, namely the context length and the option positions.

Finally, building on the causal-attention-based explanation, we design targeted interventions that can improve the performance of LLMs on QOC-structured prompts, or conversely, degrade the performance on CQO-structured prompts. These results provide additional evidence that causal attention is the driving mechanism of the sensitivity.

\section{Experimental setup}
\label{sec:setup}

\paragraph{Datasets.}
We evaluate on four reading comprehension benchmarks that require context-based reasoning: LogiQA~\citep{liu2020logiqa}; RACE-H/M~\citep{lai2017race}; SciQ~\citep{welbl2017crowdsourcing}. All tasks are in MCQA format with four options.

\paragraph{Models.}
We conduct experiments on 21 decoder-only LLMs from four model families: LLaMA 3~\citep{grattafiori2024llama}, Qwen 2.5/3~\citep{qwen2025qwen25technicalreport}, and Gemma 2~\citep{gemma_2024}. Model sizes range from 0.5B to 9B parameters, including both base and instruction-tuned variants. For architecture comparison experiments (\S\ref{sec:analysis_3}), we additionally test Flan-T5~\citep{chung2024scaling} family (encoder-decoder) and BERT, RoBERTa, ALBERT~\citep{devlin2019bert} family (encoder-only).

\paragraph{Evaluation metrics.}
We measure accuracy under two prompt orderings: CQO and QOC, and utilize the \textit{performance gap} ($\Delta = \text{Acc}_{\text{CQO}} - \text{Acc}_{\text{QOC}}$), which quantifies context order sensitivity. 

\paragraph{Other details.}
We score each architecture with the protocol that best matches its generative interface: decoder-only models are evaluated by \emph{constrained likelihood} over the four option tokens at the next-token position; encoder-only (MLM) models predict the masked answer token in a cloze template; encoder--decoder models feed the full prompt to the encoder and score decoder probabilities of the option tokens; and all generative-mode experiments (CoT, open-domain QA, RAG) parse the free-form answer from the generated text. Full templates, token sets, decoding settings, and dataset statistics are provided in Appendix~\ref{app:details}.

\section{Hypotheses and analysis}
As shown in \Cref{fig:fig2}, permuting the prompt order leads to a sharp performance drop across all benchmarks. To understand the cause, we formulate and test a set of hypotheses. All detailed experiments for each model and dataset are in~\Cref{app:additional_results}.

\begin{figure}[t]
\centering
  \begin{subfigure}{0.49\linewidth}
    \centering
    \includegraphics[width=\linewidth]{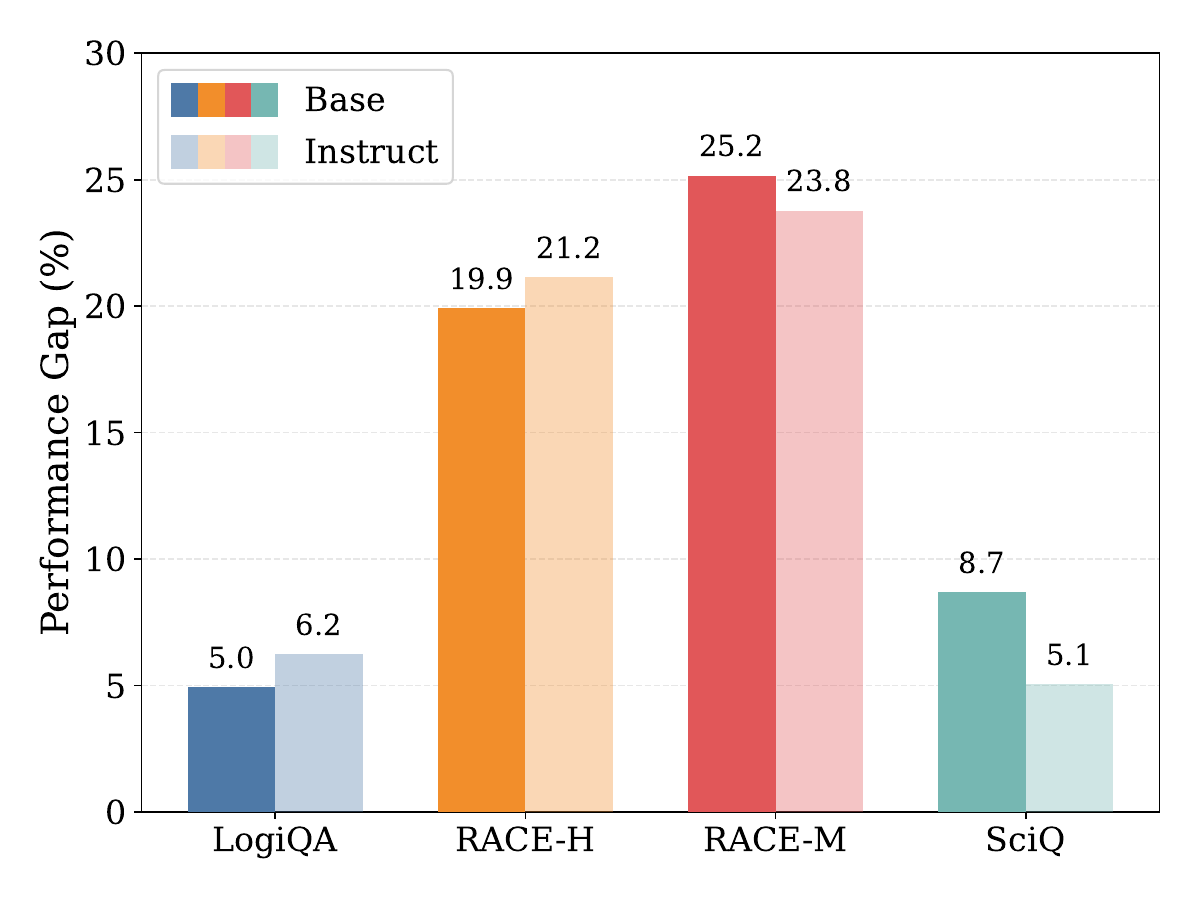}
    \subcaption{Performance gap}\label{fig:fig3-a}
  \end{subfigure}\hfill
  \begin{subfigure}{0.49\linewidth}
    \centering
    \includegraphics[width=\linewidth]{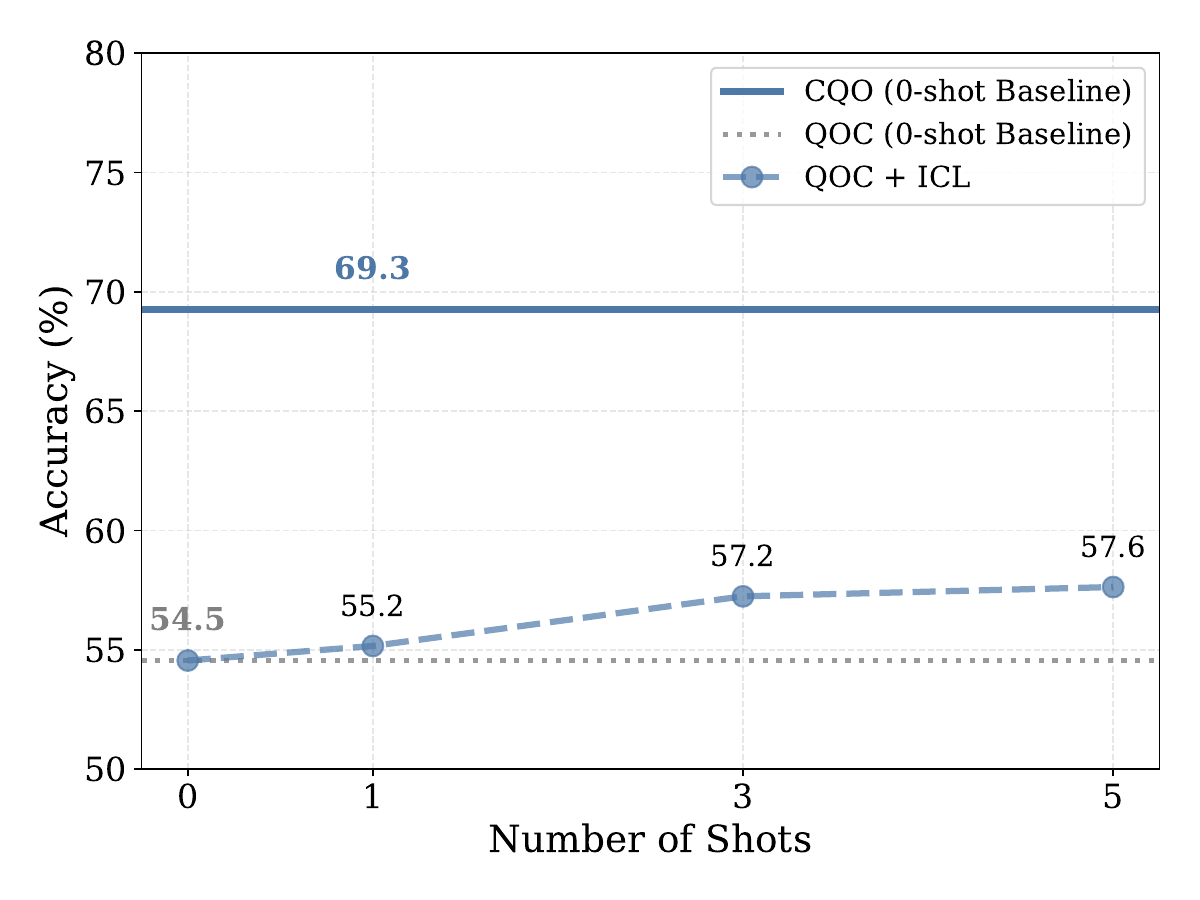}
    \subcaption{ICL performance}\label{fig:fig3-b}
  \end{subfigure}
  
  \caption{\textbf{Hypothesis 1: Effect of instruction tuning and in-context learning.} (a) We compare the \textit{performance gaps} between base and instruct models, and find they remains remarkably consistent. (b) Few-shot prompting yields marginal gains. These results suggest that the friendly formatting is not the primary driver.}
  \label{fig:analysis-tuning-icl}
\end{figure}

\subsection{Hypothesis 1: Bias in training samples}
\label{sec:analysis_1}

\textit{\textbf{Hypothesis.}} CQO is simply more frequent in the training data than QOC. Thus, in the QOC format, the model may fail to process an unfamiliar format.

\noindent\textit{\textbf{Experiment.}}
We test the training-distribution hypothesis in two ways.
First, CQO-like prompts are more common in instruction data; this hypothesis predicts a larger CQO-QOC gap for instruction models.
We therefore compare nine matched base-instruct pairs.
Second, we use in-context learning to familiarize models with the QOC format, varying the number of demonstrations up to 5-shot.

\noindent\textit{\textbf{Results.}}
As shown in \Cref{fig:fig3-a}, the CQO-QOC gaps are nearly identical, suggesting the phenomenon is not driven by instruction tuning.
Moreover, even with 5-shot demonstrations shown in \Cref{fig:fig3-b}, QOC accuracy improves by only 3.1\% and remains far below CQO.
Together, these results rule out training distribution as the primary cause.


\begin{figure}[t]
\centering
\includegraphics[width=0.8\linewidth]{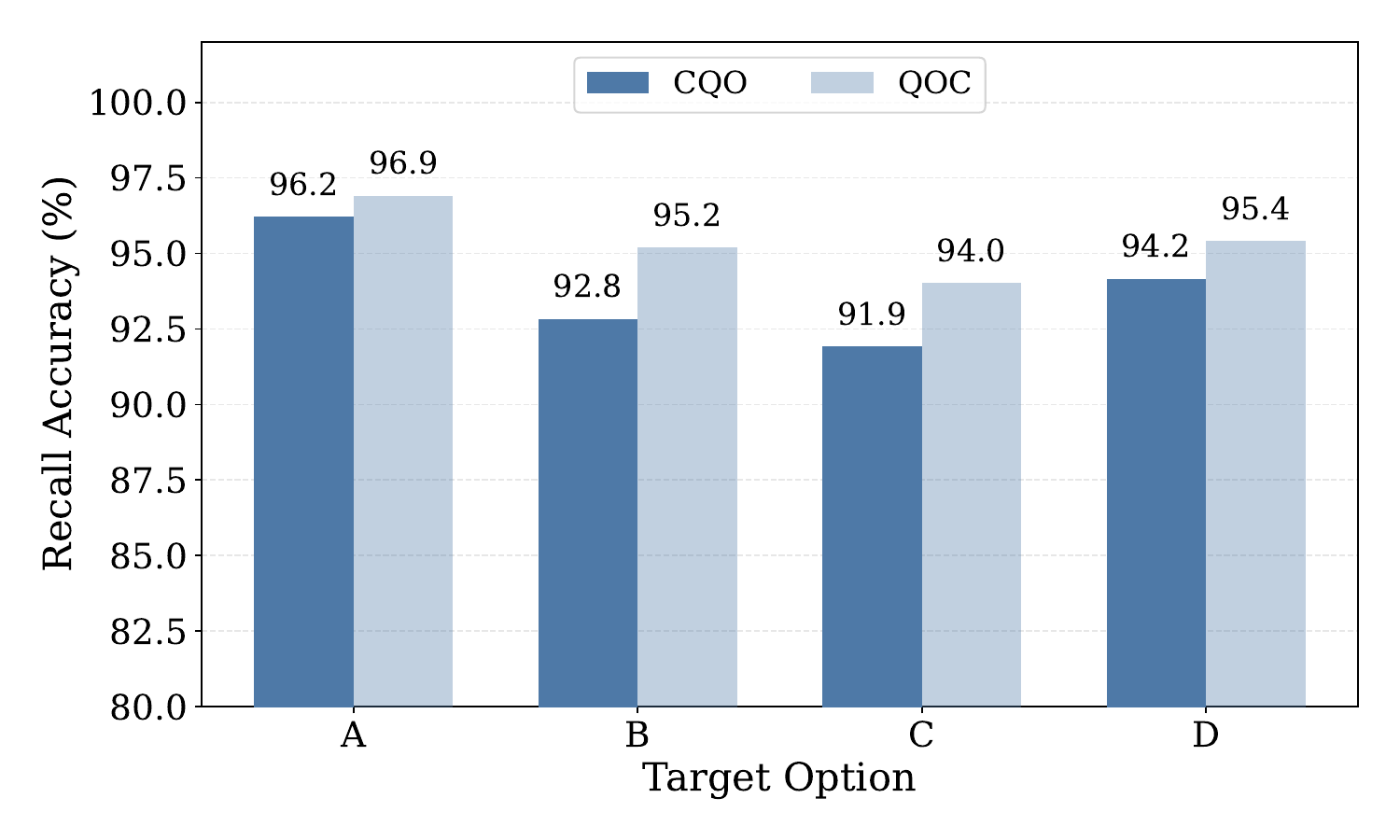}
\caption{\textbf{Hypothesis 2: Option recall analysis.} To investigate "forgetting" the options due to the long context intervention, we evaluated option recall accuracy. Results show that accuracy is consistently higher than CQO, indicating that the models retain option information and ruling out memory loss as the primary cause.}
\label{fig:fig4}
\end{figure}

\subsection{Hypothesis 2: Failure to recall options}
\label{sec:analysis_2}

\textit{\textbf{Hypothesis.}} LLMs often fail to recall the information located in the middle of the context \citep{liu2024lost}. Thus, in the QOC format, the model may fail to correctly recall the options, which are located between the question and the context.

\noindent\textit{\textbf{Experiment.}} After presenting the prompt, we ask the LLMs to recall each option. Precisely, we measure the chance of an exact match for each option.

\noindent\textit{\textbf{Results.}} As shown in \Cref{fig:fig4}, QOC achieves a similar, or even higher, recall accuracy than CQO. This indicates that the failure of option recall may not be the cause of accuracy drops in QOC.


\subsection{Hypothesis 3: Causal attention}
\label{sec:analysis_3}

\textit{\textbf{Hypothesis.}} Causal attention mask prevents option tokens from attending to context in QOC. In decoder-only, each token can only attend to preceding tokens. Thus, in QOC, where options appear before context, option representations are computed \emph{without} information from the context.

\noindent\textit{\textbf{Experiment 1: Architecture comparison.}}
If causal masking is the root cause, models with bidirectional attention should exhibit no \textit{performance gap}.
We compare three architecture types: decoder-only (causal), encoder-decoder (bidirectional encoder), and encoder-only (bidirectional).
For encoder-decoder models, we feed the entire prompt to the encoder and let the decoder generate the answer.

\begin{figure}[t]
\centering
  \begin{subfigure}{0.48\linewidth}
    \centering
    \includegraphics[width=\linewidth]{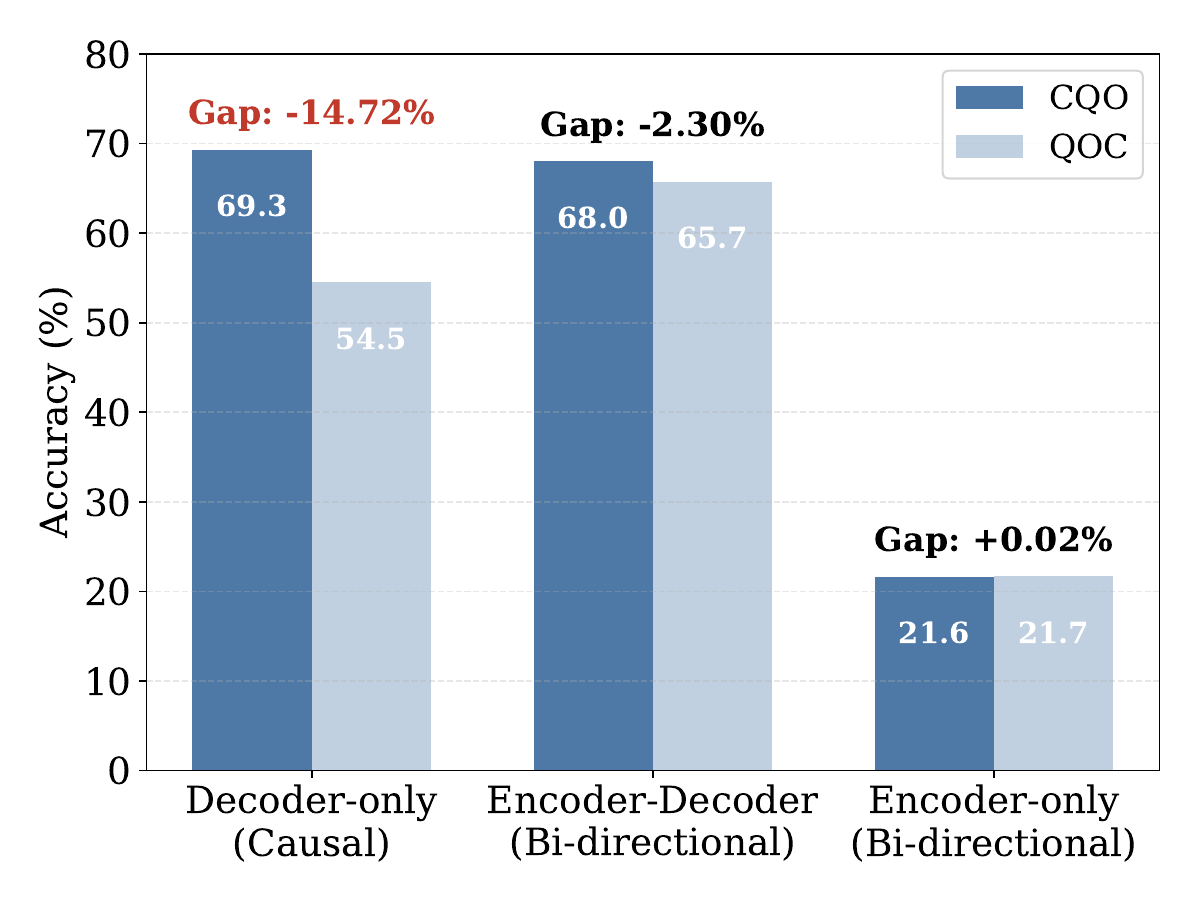}
    \subcaption{Model design comparison}\label{fig:fig5-a}
  \end{subfigure}\hfill
  \begin{subfigure}{0.48\linewidth}
    \centering
    \includegraphics[width=\linewidth]{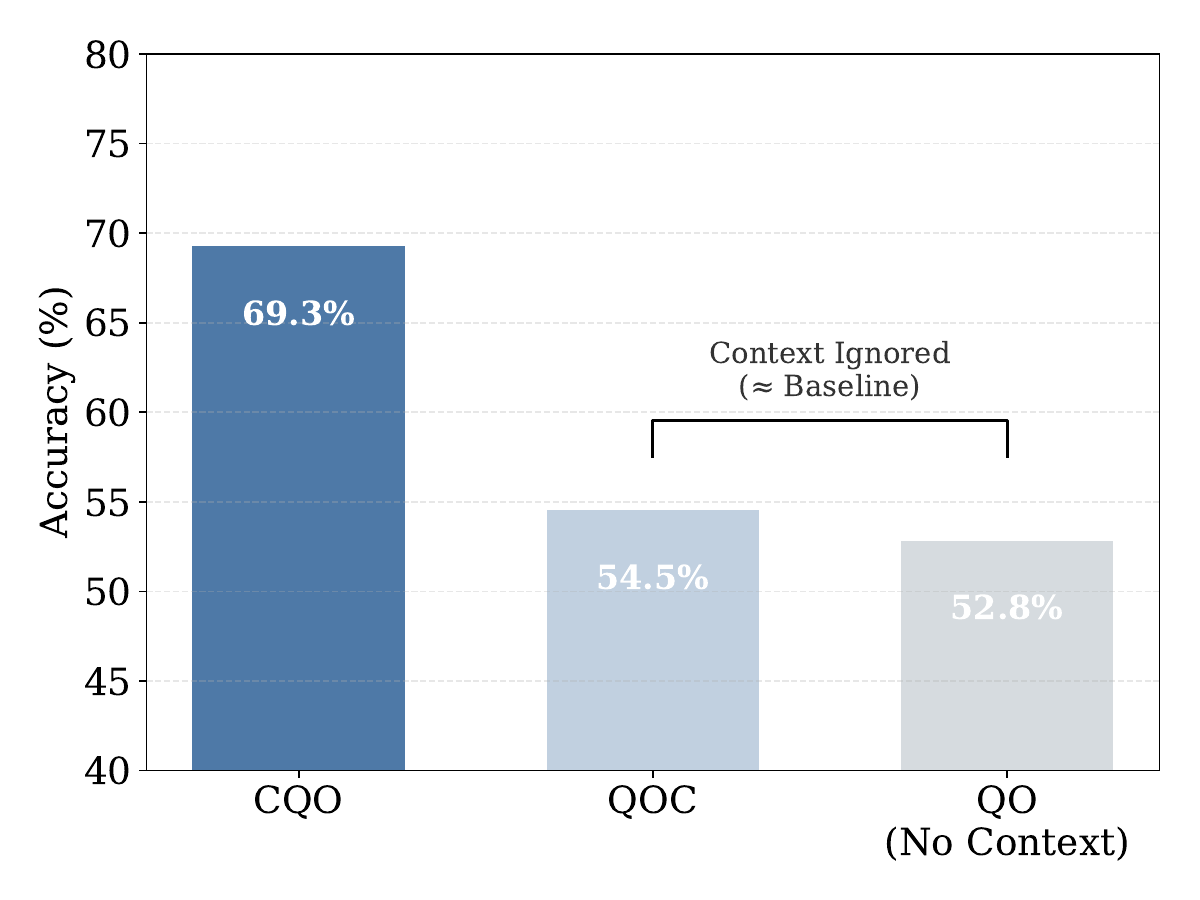}
    \subcaption{Context position}\label{fig:fig5-b}
  \end{subfigure}
  
  \caption{\textbf{Hypothesis 3, Exp\# 1: Architecture comparison.} (a) Decoder-only LLMs show a large gap. In contrast, encoder-only and encoder-decoder LLMs have a minimal accuracy gap, confirming that the causal mask is the primary factor. (b) For decoder-only LLMs, QOC performance drops to nearly QO, indicating that the information inside the context is ignored.}
  \label{fig:causal-mechanism}
\end{figure}

\begin{figure}[t]
\centering
  \begin{subfigure}{0.48\linewidth}
    \centering
    \includegraphics[width=\linewidth]{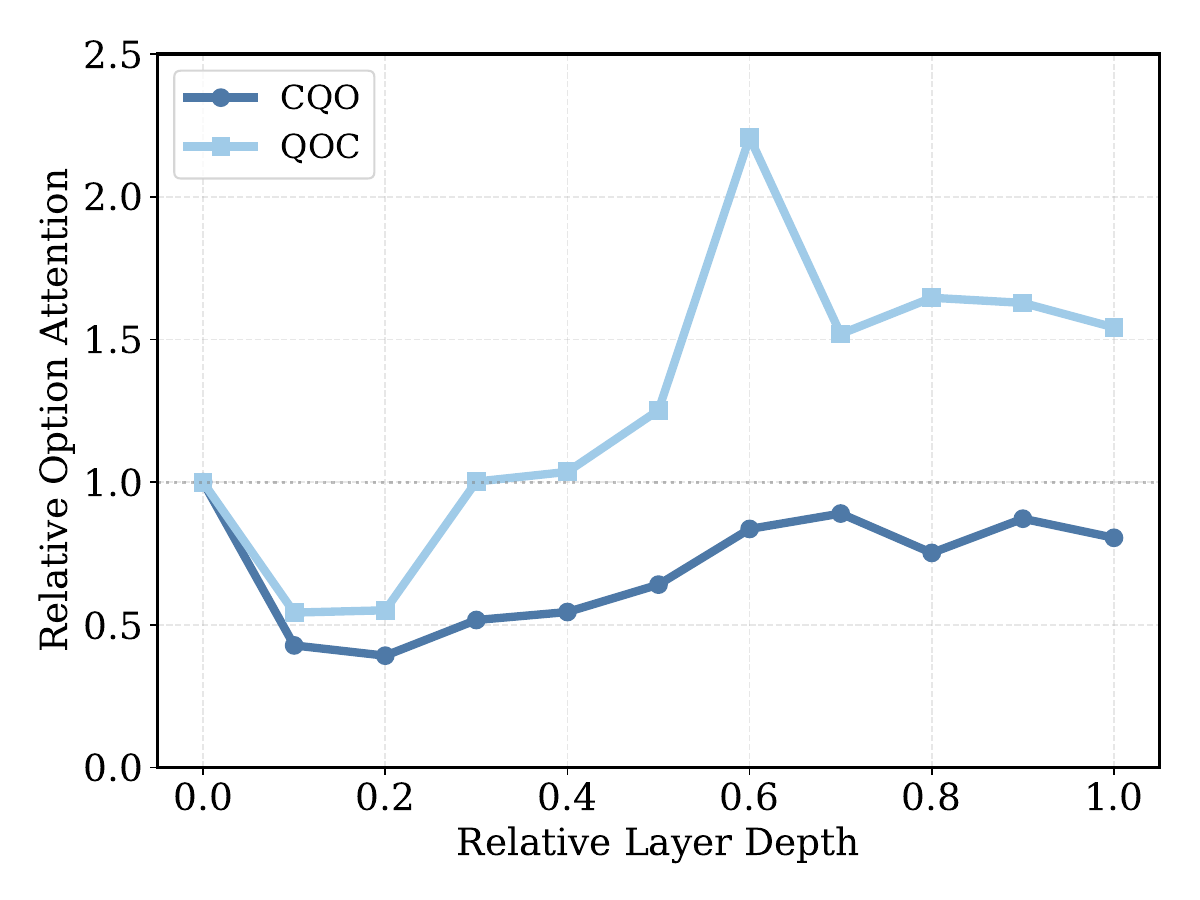}
    \subcaption{Attention decay}\label{fig:fig6-a}
  \end{subfigure}\hfill
  \begin{subfigure}{0.48\linewidth}
    \centering
    \includegraphics[width=\linewidth]{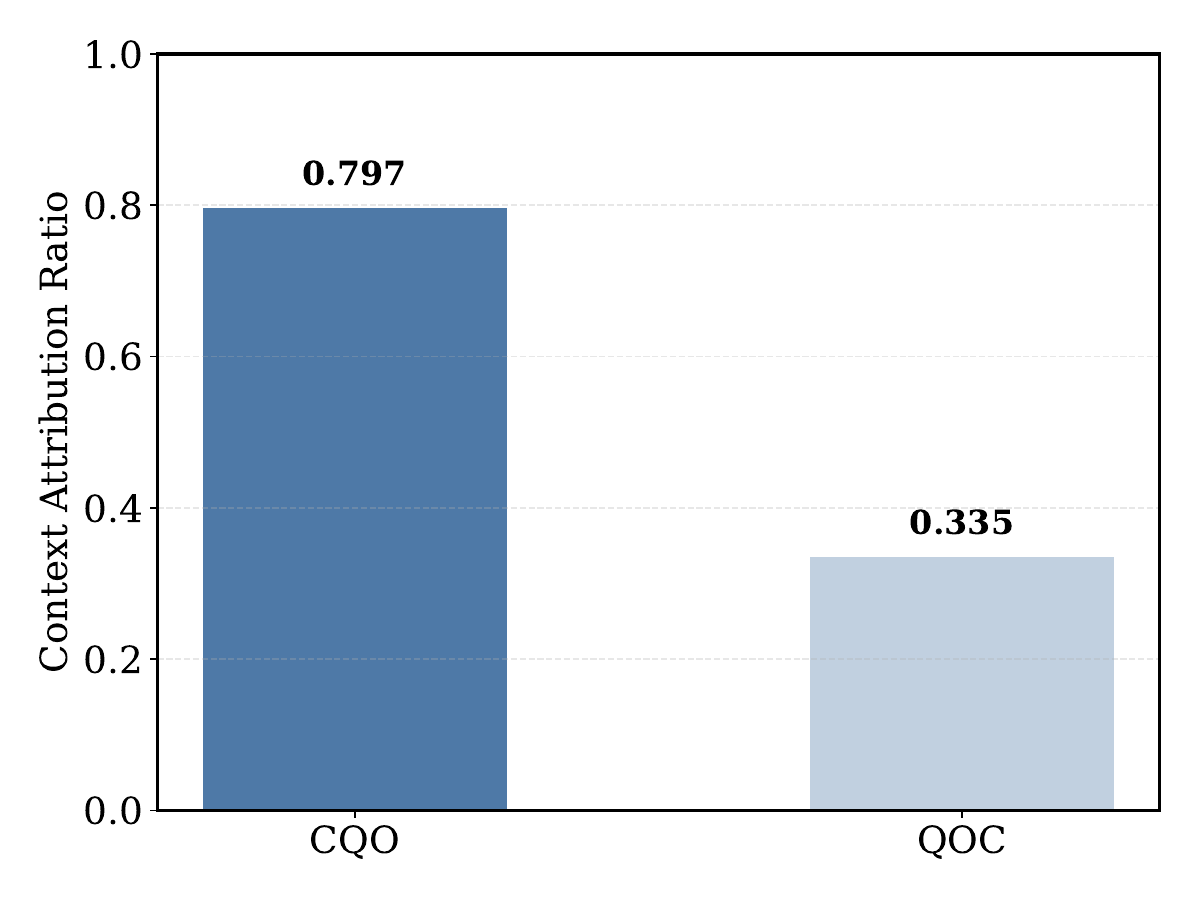}
    \subcaption{Context attribution ratio}\label{fig:fig6-b}
  \end{subfigure}
  
  \caption{\textbf{Hypothesis 3, Exp\# 2 \& 3: Analysis of context utilization.} (a) Layer-wise option attention: In CQO, attention to options declines as context is integrated, whereas in QOC it rises. (b) Gradient-based attribution: context tokens contribute substantially more to CQO predictions than to QOC.}
  \label{fig:attention-attribution}
\end{figure}

\noindent\textit{\textbf{Results 1.}}
As shown in \Cref{fig:fig5-a}, decoder-only models show a 14.72\% gap, encoder-decoder models (Flan-T5) show 2.30\%, and encoder-only models (BERT, RoBERTa, ALBERT) show a near-zero gap (0.02\%).
This pattern strongly implicates causal masking as the underlying mechanism.
Results on additional architectures (state-space models, hybrid linear attention, and diffusion language models) are reported in Appendix~\ref{app:other_arch}.

\noindent\textit{\textbf{Experiment 2: Context removal test.}}
If the context is effectively inaccessible in QOC, then removing it entirely should not materially change performance.
We therefore compare QOC against QO (Question-Options only), where the context is omitted.

\noindent\textit{\textbf{Results 2.}}
As shown in \Cref{fig:fig5-b}, QOC accuracy is nearly identical to QO, consistent with the model failing to use the context in QOC.
This supports the hypothesis that the issue is not forgetting but restricted access induced by the causal mask.

\noindent\textit{\textbf{Experiment 3: Attention analysis.}}
We analyze how attention to option tokens evolves across layers.
Also, we measure token contributions using Gradient$\times$Input attribution~\citep{shrikumar2017learning}, aggregating per-token scores over the context span and normalizing by total attribution.

\noindent\textit{\textbf{Results 3.}}
Option tokens receive exactly zero attention from context tokens in QOC by construction of the causal mask, confirming that the context-to-option pathway is blocked.
Across layers, attention to options increases with depth in QOC but decreases in CQO, suggesting that QOC increasingly relies on the options themselves while CQO more effectively integrates contextual information (see~\Cref{fig:fig6-a}).
Also, context attribution is 0.797 in CQO but only 0.335 in QOC (see~\Cref{fig:fig6-b}).

\paragraph{Why doesn't the final attention fix it?}
One may argue that the final answer token can still attend to the context in QOC. However, causal masking prevents the options themselves from being context-conditioned: all option hidden states are computed before any context is seen, so they cannot encode evidence--option alignment. Since later tokens cannot retroactively update earlier states (they can only read them), the model must do evidence--option comparison only at the final decoding step using context-blind option representations.
This single-step-bottleneck view makes a concrete, testable prediction: interventions that either re-expose option tokens to the context (option repetition) or grant the model extra decoding steps past the full prompt (CoT prompting) should partially close the CQO--QOC gap, which is exactly what we observe in~\Cref{sec:interventions}. A formal information-theoretic derivation, showing that under QOC the mutual information between option representations and the context is zero by construction, is given in Appendix~\ref{app:theory}.

\paragraph{Modulating factors.} Two factors modulate the gap severity. First, longer contexts show larger gaps (\Cref{tab:context_gap}): more context means more information becomes inaccessible under causal masking. 
Second, earlier answer positions suffer greater accuracy drops than later ones in QOC (\Cref{tab:position_gap}), as later options are near the context.

  

\begin{table}[t]
    \centering
    \captionsetup[subtable]{justification=centering} 

    \begin{subtable}[t]{\linewidth}
        \centering
    
        \resizebox{0.75\linewidth}{!}{%
        \begin{tabular}{lcccc}
            \toprule
            & \multicolumn{2}{c}{\textbf{Short context}} & \multicolumn{2}{c}{\textbf{Long context}} \\
            \cmidrule(lr){2-3} \cmidrule(lr){4-5}
            \textbf{Dataset} & LogiQA & SciQ & RACE-M & RACE-H \\
            \midrule
            \textbf{Avg. Length} & $\sim$70 & $\sim$70 & $\sim$195 & $\sim$305 \\
            \textbf{Gap ($\Delta$)} & 6.2\% & 7.3\% & 24.8\% & 20.8\% \\
            \bottomrule
        \end{tabular}%
        }
        \caption{\textbf{Effect of context length.}}
        \label{tab:context_gap}
    \end{subtable}

    \bigskip 

    \begin{subtable}[t]{\linewidth}
        \centering
        
        \resizebox{0.7\linewidth}{!}{%
        \begin{tabular}{lcccc}
            \toprule
            \textbf{Position} & A & B & C & \textbf{D} \\
            \midrule
            \textbf{Gap ($\Delta$)} & 22.4\% & 20.5\% & 19.2\% & \textbf{9.9\%} \\
            \bottomrule
        \end{tabular}%
        }
        \caption{\textbf{Effect of answer position.}}
        \label{tab:position_gap}
    \end{subtable}

    \caption{\textbf{Factors influencing causal masking impact.}
    (a) Datasets with longer contexts suffer more from causal masking.
    (b) The correct answer's position affects the gap, with option D being the most robust.}
    \label{tab:combined_analysis}
\end{table}

\section{Targeted interventions}
\label{sec:interventions}

We design targeted interventions that directly manipulate the option and context attention pathway (see \Cref{tab:restoration_results}). We apply each intervention to all 21 decoder-only models across four datasets.

\paragraph{Attention pruning (degrading CQO).}
To simulate the constraint imposed by causal masking in QOC, we block option-to-context attention in CQO. For each sample, we identify token spans for the context and the options, and set $\text{mask}[i,j] = -\infty$ for all pairs where $i \in \text{Options}$ and $j \in \text{Context}$. This prevents option tokens from attending to context tokens while leaving all other attention unchanged. CQO accuracy drops from 69.26\% to 42.46\% ($-26.8$), with consistent decreases across model families (Qwen: $28.1$\%, LLaMA: $25.3$\%, Gemma: $26.9$\%), indicating that option access to contextual evidence is critical.

\paragraph{Activation patching (improving QOC).}
We restore context-aware option representations in QOC by patching option hidden states with those computed under CQO. For each sample, we run both templates and align corresponding option token positions by exact string matching. At layers in the middle-to-late half of the network (e.g., layers 12--23 for 24-layer models, normalized by depth), we replace $h^{\text{QOC}}_{\text{opt}}$ with $h^{\text{CQO}}_{\text{opt}}$. We patch only option tokens (not context or question) to isolate the mechanism; patched states come from a different template on the same sample, and token alignment is verified by exact string match. This increases QOC accuracy by $6.0$ points on average, with larger gains for models exhibiting larger baseline gaps.

\paragraph{Option repetition (improving QOC).}
As a simple prompt, we repeat the options after the context (QOCO). The repeated option tokens can attend to the context under the causal mask, requiring no internal model intervention. Repeating options improves QOC by $8.2$ points, partially closing the gap without modifying model internals.

\paragraph{Chain-of-thought prompting (improving QOC).}
A natural follow-up is whether extra decoding steps can lift the bottleneck: under chain-of-thought (CoT) prompting, newly generated reasoning tokens lie after the full prompt and may attend to both the options \emph{and} the context, producing context-aware option representations on the fly. Replacing greedy answer scoring with ``\textit{Let's think step by step}'' CoT shrinks the CQO--QOC gap from $14.72$ to $7.47$ on average (see~\Cref{tab:restoration_results}), without any architecture change. This is consistent with the causal-mask account: the remaining residual gap reflects the fact that option tokens are still initially encoded without context, and only a few reasoning steps are available to recover the missing evidence--option alignment. Results on a larger closed-source model (Gemini), where CoT reduces the gap by $83.4\%$, are reported in Appendix~\ref{app:closed_source}.


\begin{table}[t]
\centering
\resizebox{\linewidth}{!}{%
\begin{tabular}{lcccc|c}
\toprule
\textbf{Methods} & \textbf{LogiQA} & \textbf{SciQ} & \textbf{RACE-M} & \textbf{RACE-H} & \textbf{Avg} \\
\midrule
CQO & 39.08 & 94.16 & 74.32 & 69.48 & 69.26 \\
\quad + Attention pruning & 30.47 & 60.94 & 39.52 & 38.89 & 42.46 \textcolor{red}{(-26.8)} \\
\midrule
QOC & 32.94 & 86.89 & 49.57 & 48.76 & 54.54 \\
\quad + Activation patching & 35.18 & 87.64 & 62.22 & 56.94 & 60.49 \textcolor{blue}{(+6.0)} \\
\quad + QOCO (Repeat) & 35.62 & 91.80 & 63.55 & 59.65 & 62.76 \textcolor{blue}{(+8.2)} \\
\midrule
CQO-CoT & 32.79 & 78.88 & 63.33 & 57.44 & 58.11 \\
QOC-CoT & 29.47 & 73.62 & 51.99 & 47.49 & 50.64 \\
\quad Gap (CQO-CoT $-$ QOC-CoT) & 3.32 & 5.26 & 11.34 & 9.95 & 7.47 \\
\bottomrule
\end{tabular}
}
\caption{\textbf{Targeted interventions close the CQO--QOC gap.} Accuracies are averaged over 21 decoder-only models. The upper block uses logit-based scoring; the lower CoT block uses generative scoring and is not directly comparable in absolute value---what matters is the gap shrinking from 14.72 to 7.47 ($-7.25$), corroborating that the bottleneck is about option representations, not downstream reasoning.}
\label{tab:restoration_results}
\end{table}

\section{Conclusion}

We investigated why decoder-only LLMs exhibit a huge performance gap between CQO and QOC orderings in MCQA tasks. After ruling out training distribution bias and memory decay, we identified causal attention as the core mechanism: in QOC, the causal mask prevents options from attending to context, rendering it effectively inaccessible. This finding is supported by architecture comparisons (encoder models show no gap), attention analysis, and targeted interventions that successfully degrade CQO or improve QOC performance. Our work provides mechanistic insight into prompt sensitivity and offers practical guidance.

\section*{Limitations}
Our work has two main limitations.
First, the theoretical account of the single-step bottleneck in Appendix~\ref{app:theory} is deliberately minimal: it establishes that option hidden states under QOC are structurally independent of the context ($I(h_O^{\mathrm{QOC}}; C \mid Q, O) = 0$) and that interventions which lift this constraint (option repetition, activation patching, CoT) close a large fraction of the gap.
Second, this paper is diagnostic in scope: we identify \emph{why} decoder-only LLMs suffer from prompt-order sensitivity and demonstrate several inference-time mitigations (CoT, QOCO, activation patching), but we do not yet propose a practical, training-time fix that closes the gap without any runtime overhead. We leave for future work.

\section*{Acknowledgments}
This work was supported in part by the Institute of Information \& Communications Technology Planning \& Evaluation (IITP) grant funded by the Korea government (MSIT) (Nos. RS-2024-00457882, No.RS-2019-II191906) and in part by the National Research Foundation of Korea (NRF) grant funded by the Korea government (MSIT) (Nos. RS-2024-00453301, RS-2025-24873016, RS-2026-25494004).

\bibliography{custom}

@article{wei2022chain,
  title={Chain-of-thought prompting elicits reasoning in large language models},
  author={Wei, Jason and Wang, Xuezhi and Schuurmans, Dale and Bosma, Maarten and Xia, Fei and Chi, Ed and Le, Quoc V and Zhou, Denny and others},
  journal={Advances in neural information processing systems},
  volume={35},
  pages={24824--24837},
  year={2022}
}

@inproceedings{shrikumar2017learning,
  title={Learning important features through propagating activation differences},
  author={Shrikumar, Avanti and Greenside, Peyton and Kundaje, Anshul},
  booktitle={International conference on machine learning},
  pages={3145--3153},
  year={2017},
  organization={PMlR}
}

@article{liu2024lost,
  title={Lost in the middle: How language models use long contexts},
  author={Liu, Nelson F and Lin, Kevin and Hewitt, John and Paranjape, Ashwin and Bevilacqua, Michele and Petroni, Fabio and Liang, Percy},
  journal={Transactions of the Association for Computational Linguistics},
  volume={12},
  pages={157--173},
  year={2024}
}

@inproceedings{li2024loogle,
  title={Loogle: Can long-context language models understand long contexts?},
  author={Li, Jiaqi and Wang, Mengmeng and Zheng, Zilong and Zhang, Muhan},
  booktitle={Proceedings of the 62nd Annual Meeting of the Association for Computational Linguistics (Volume 1: Long Papers)},
  pages={16304--16333},
  year={2024}
}

@article{an2024make,
  title={Make your llm fully utilize the context},
  author={An, Shengnan and Ma, Zexiong and Lin, Zeqi and Zheng, Nanning and Lou, Jian-Guang and Chen, Weizhu},
  journal={Advances in Neural Information Processing Systems},
  volume={37},
  pages={62160--62188},
  year={2024}
}

@inproceedings{meng2022locating,
  title={Locating and Editing Factual Associations in {GPT}},
  author={Meng, Kevin and Bau, David and Andonian, Alex and Belinkov, Yonatan},
  booktitle={Advances in Neural Information Processing Systems},
  volume={35},
  pages={17359--17372},
  year={2022}
}

@article{vig2020investigating,
  title={Investigating gender bias in language models using causal mediation analysis},
  author={Vig, Jesse and Gehrmann, Sebastian and Belinkov, Yonatan and Qian, Sharon and Nevo, Daniel and Singer, Yaron and Shieber, Stuart},
  journal={Advances in neural information processing systems},
  volume={33},
  pages={12388--12401},
  year={2020}
}

@inproceedings{ding2021evaluating,
  title={Evaluating Saliency Methods for Neural Language Models},
  author={Ding, Shuoyang and Koehn, Philipp},
  booktitle={Proceedings of the 2021 Conference of the North American Chapter of the Association for Computational Linguistics: Human Language Technologies},
  pages={5034--5052},
  year={2021}
}

@inproceedings{poche-etal-2025-consim,
    title = "{C}on{S}im: Measuring Concept-Based Explanations' Effectiveness with Automated Simulatability",
    author = "Poch{\'e}, Antonin  and
      Jacovi, Alon  and
      Picard, Agustin Martin  and
      Boutin, Victor  and
      Jourdan, Fanny",
    editor = "Che, Wanxiang  and
      Nabende, Joyce  and
      Shutova, Ekaterina  and
      Pilehvar, Mohammad Taher",
    booktitle = "Proceedings of the 63rd Annual Meeting of the Association for Computational Linguistics (Volume 1: Long Papers)",
    month = jul,
    year = "2025",
    address = "Vienna, Austria",
    publisher = "Association for Computational Linguistics",
    url = "https://aclanthology.org/2025.acl-long.279/",
    doi = "10.18653/v1/2025.acl-long.279",
    pages = "5594--5615",
    ISBN = "979-8-89176-251-0"
}

@inproceedings{abnar2020quantifying,
  title={Quantifying Attention Flow in Transformers},
  author={Abnar, Samira and Zuidema, Willem},
  booktitle={Proceedings of the 58th Annual Meeting of the Association for Computational Linguistics},
  pages={4190--4197},
  year={2020}
}

@inproceedings{clark2019does,
  title={What Does BERT Look at? An Analysis of BERT’s Attention},
  author={Clark, Kevin and Khandelwal, Urvashi and Levy, Omer and Manning, Christopher D},
  booktitle={Proceedings of the 2019 ACL Workshop BlackboxNLP: Analyzing and Interpreting Neural Networks for NLP},
  pages={276--286},
  year={2019}
}

@article{geiger2021causal,
  title={Causal abstractions of neural networks},
  author={Geiger, Atticus and Lu, Hanson and Icard, Thomas and Potts, Christopher},
  journal={Advances in Neural Information Processing Systems},
  volume={34},
  pages={9574--9586},
  year={2021}
}

@inproceedings{lu2022fantastically,
  title={Fantastically ordered prompts and where to find them: Overcoming few-shot prompt order sensitivity},
  author={Lu, Yao and Bartolo, Max and Moore, Alastair and Riedel, Sebastian and Stenetorp, Pontus},
  booktitle={Proceedings of the 60th Annual Meeting of the Association for Computational Linguistics (Volume 1: Long Papers)},
  pages={8086--8098},
  year={2022}
}

@inproceedings{pezeshkpour-hruschka-2024-large,
    title = "Large Language Models Sensitivity to The Order of Options in Multiple-Choice Questions",
    author = "Pezeshkpour, Pouya  and
      Hruschka, Estevam",
    editor = "Duh, Kevin  and
      Gomez, Helena  and
      Bethard, Steven",
    booktitle = "Findings of the Association for Computational Linguistics: NAACL 2024",
    month = jun,
    year = "2024",
    address = "Mexico City, Mexico",
    publisher = "Association for Computational Linguistics",
    url = "https://aclanthology.org/2024.findings-naacl.130/",
    doi = "10.18653/v1/2024.findings-naacl.130",
    pages = "2006--2017"
}

@inproceedings{
zheng2024large,
title={Large Language Models Are Not Robust Multiple Choice Selectors},
author={Chujie Zheng and Hao Zhou and Fandong Meng and Jie Zhou and Minlie Huang},
booktitle={The Twelfth International Conference on Learning Representations},
year={2024},
url={https://openreview.net/forum?id=shr9PXz7T0}
}

@inproceedings{
sclar2024quantifying,
title={Quantifying Language Models' Sensitivity to Spurious Features in Prompt Design or: How I learned to start worrying about prompt formatting},
author={Melanie Sclar and Yejin Choi and Yulia Tsvetkov and Alane Suhr},
booktitle={The Twelfth International Conference on Learning Representations},
year={2024},
url={https://openreview.net/forum?id=RIu5lyNXjT}
}

@inproceedings{
wiegreffe2025answer,
title={Answer, Assemble, Ace: Understanding How {LM}s Answer Multiple Choice Questions},
author={Sarah Wiegreffe and Oyvind Tafjord and Yonatan Belinkov and Hannaneh Hajishirzi and Ashish Sabharwal},
booktitle={The Thirteenth International Conference on Learning Representations},
year={2025},
url={https://openreview.net/forum?id=6NNA0MxhCH}
}

@inproceedings{shaier2024not,
  title={It is not about what you say, it is about how you say it: A surprisingly simple approach for improving reading comprehension},
  author={Shaier, Sagi and Hunter, Lawrence and Wense, Katharina},
  booktitle={Findings of the Association for Computational Linguistics: ACL 2024},
  pages={8292--8305},
  year={2024}
}

@inproceedings{kojima2022large,
  title={Large language models are zero-shot reasoners},
  author={Kojima, Takeshi and Gu, Shixiang Shane and Reid, Machel and Matsuo, Yutaka and Iwasawa, Yusuke},
  booktitle={Advances in Neural Information Processing Systems},
  volume={35},
  pages={22199--22213},
  year={2022}
}

@inproceedings{liu2020logiqa,
  title={LogiQA: A Challenge Dataset for Machine Reading Comprehension with Logical Reasoning},
  author={Liu, Jian and Cui, Leyang and Liu, Hanmeng and Huang, Dandan and Wang, Yile and Zhang, Yue},
  booktitle={Proceedings of the Twenty-Ninth International Joint Conference on Artificial Intelligence (IJCAI)},
  pages={3622--3628},
  year={2020}
}

@inproceedings{lai2017race,
  title={{RACE}: Large-scale ReAding Comprehension Dataset From Examinations},
  author={Lai, Guokun and Xie, Qizhe and Liu, Hanxiao and Yang, Yiming and Hovy, Eduard},
  booktitle={Proceedings of the 2017 Conference on Empirical Methods in Natural Language Processing (EMNLP)},
  pages={785--794},
  year={2017}
}

@inproceedings{welbl2017crowdsourcing,
  title={Crowdsourcing Multiple Choice Science Questions},
  author={Welbl, Johannes and Liu, Nelson F. and Gardner, Matt},
  booktitle={Proceedings of the 3rd Workshop on Noisy User-generated Text},
  pages={94--106},
  year={2017}
}

@article{grattafiori2024llama,
  title={The Llama 3 Herd of Models},
  author={Grattafiori, Aaron and Dubey, Abhimanyu and Jauhri, Abhinav and others},
  journal={arXiv preprint arXiv:2407.21783},
  year={2024}
}

@article{qwen2025qwen25technicalreport,
  title={Qwen2. 5 Technical Report},
  author={Yang, An and Yang, Baosong and Zhang, Beichen and Hui, Binyuan and Zheng, Bo and Yu, Bowen and Li, Chengyuan and Liu, Dayiheng and Huang, Fei and Wei, Haoran and others},
  journal={arXiv preprint arXiv:2412.15115},
  year={2024}
}

@article{gemma_2024,
  title={Gemma 2: Improving open language models at a practical size},
  author={Team, Gemma and Riviere, Morgane and Pathak, Shreya and Sessa, Pier Giuseppe and Hardin, Cassidy and Bhupatiraju, Surya and Hussenot, L{\'e}onard and Mesnard, Thomas and Shahriari, Bobak and Ram{\'e}, Alexandre and others},
  journal={arXiv preprint arXiv:2408.00118},
  year={2024}
}

@article{chung2024scaling,
  title={Scaling instruction-finetuned language models},
  author={Chung, Hyung Won and Hou, Le and Longpre, Shayne and Zoph, Barret and Tay, Yi and Fedus, William and Li, Yunxuan and Wang, Xuezhi and Dehghani, Mostafa and Brahma, Siddhartha and others},
  journal={Journal of Machine Learning Research},
  volume={25},
  number={70},
  pages={1--53},
  year={2024}
}

@inproceedings{devlin2019bert,
  title={{BERT}: Pre-training of Deep Bidirectional Transformers for Language Understanding},
  author={Devlin, Jacob and Chang, Ming-Wei and Lee, Kenton and Toutanova, Kristina},
  booktitle={Proceedings of the 2019 Conference of the North American Chapter of the Association for Computational Linguistics (NAACL)},
  pages={4171--4186},
  year={2019}
}

\clearpage

\appendix

\section{Related work}
\label{sec:related}

\paragraph{Prompt sensitivity.}
A growing body of work has established that LLMs can be sensitive to prompt design choices, including input order and formatting.
In the in-context learning (ICL) setting, \citet{lu2022fantastically} show that reordering the same set of demonstrations can substantially change accuracy and propose strategies to mitigate order sensitivity, but do not explore the underlying internal mechanism.
In MCQA, prior work reports large variation under option permutations and develops mitigation methods (e.g., \citet{pezeshkpour-hruschka-2024-large, zheng2024large}), yet largely remains at the level of behavioral characterization.
Prompt formatting studies further show that semantically preserving changes---punctuation, spacing, labeling, or layout---can induce sizable performance swings (e.g., \citet{sclar2024quantifying}).
More mechanistic MCQA analyses, such as \citet{wiegreffe2025answer}, use activation patching-style interventions to localize where answer selection is determined and how it is amplified across layers.
However, these lines of work do not directly address a block-order question that arises with long contexts: whether and how the accessibility of context information to option representations changes with prompt order.
Relatedly, \citet{shaier2024not} reports large order effects in reading comprehension, but stops short of providing a structural account of the failure mode.
In contrast, we study a striking order effect in MCQA between \textsc{CQO} and \textsc{QOC} and identify an architectural mechanism: under causal attention, option tokens in \textsc{QOC} are structurally prevented from integrating information from the subsequent context.
We then test this mechanism with targeted causal interventions that manipulate the option--context information pathway.

\paragraph{Mechanistic analysis tools.}
To connect prompt-order sensitivity to internal computation, we draw on a set of mechanistic interpretability tools.
First, we use attention analysis to characterize how models route information under different prompt orders; prior work has shown the utility of interpreting attention weights as explanations~\citep{clark2019does, abnar2020quantifying}.
We compute attention statistics for each prompt order, use them as descriptive diagnostics, and also perform gradient-based analyses.
We employ gradient-based input attribution as an auxiliary diagnostic: Gradient$\times$Input and related attribution methods quantify how sensitive the prediction is to perturbations of each input token representation, and have been widely used as simple, interpretable feature-importance baselines~\citep{shrikumar2017learning, ding2021evaluating, poche-etal-2025-consim}.
We also use causal interventions on internal pathways. In particular, we (i) ablate the option-to-context attention pathway by masking attention edges between token groups (a targeted path/edge ablation), and (ii) apply activation patching (causal tracing/interchange interventions), which swaps hidden states from a ``clean'' run into a ``corrupted'' run to test whether a specific representation is causally responsible for the downstream behavior~\citep{vig2020investigating, geiger2021causal, meng2022locating}.
We adapt these tools to the prompt-order setting by isolating the option--context information pathway: attention ablation tests necessity, and activation patching tests sufficiency of context-conditioned option representations for closing the CQO-QOC gap.

\paragraph{Long-context failure modes.}
Prior work shows that LLMs can underutilize long contexts and are sensitive to where relevant evidence appears (e.g., lost-in-the-middle; \citet{liu2024lost,li2024loogle,an2024make}). In our setting, we use long contexts to test a specific hypothesis: QOC may underperform because the model forgets the options after reading a long context. Our option-recall results reject this explanation---QOC retains options at least as well as CQO---suggesting that the CQO--QOC gap is not primarily a long-context memory failure, but a distinct prompt-order effect.


\section{Detailed information}
\label{app:details}

To demonstrate the robustness of our findings, we conduct experiments with various models and datasets. Details are below.

\subsection{Model information}
\label{app:models}

\begin{table*}[t]
\small
\centering
\vspace{0.25cm}
\resizebox{0.65\textwidth}{!}{
\begin{tabular}{ll}
\toprule
\textbf{Model} & \textbf{Hugging Face ID} \\
\midrule
\multicolumn{2}{l}{\textbf{Qwen 2.5 Family}} \\
\midrule
Qwen2.5-0.5B            & \texttt{Qwen/Qwen2.5-0.5B} \\
Qwen2.5-0.5B-Instruct   & \texttt{Qwen/Qwen2.5-0.5B-Instruct} \\
Qwen2.5-1.5B            & \texttt{Qwen/Qwen2.5-1.5B} \\
Qwen2.5-1.5B-Instruct   & \texttt{Qwen/Qwen2.5-1.5B-Instruct} \\
Qwen2.5-3B-Instruct     & \texttt{Qwen/Qwen2.5-3B-Instruct} \\
Qwen2.5-7B              & \texttt{Qwen/Qwen2.5-7B} \\
Qwen2.5-7B-Instruct     & \texttt{Qwen/Qwen2.5-7B-Instruct} \\
\midrule
\multicolumn{2}{l}{\textbf{Qwen 3 Family}} \\
\midrule
Qwen3-0.6B              & \texttt{Qwen/Qwen3-0.6B} \\
Qwen3-1.7B              & \texttt{Qwen/Qwen3-1.7B} \\
Qwen3-4B                & \texttt{Qwen/Qwen3-4B} \\
\midrule
\multicolumn{2}{l}{\textbf{LLaMA 3 Family}} \\
\midrule
Llama-3.2-1B            & \texttt{meta-llama/Llama-3.2-1B} \\
Llama-3.2-1B-Instruct   & \texttt{meta-llama/Llama-3.2-1B-Instruct} \\
Llama-3.2-3B            & \texttt{meta-llama/Llama-3.2-3B} \\
Llama-3.2-3B-Instruct   & \texttt{meta-llama/Llama-3.2-3B-Instruct} \\
Llama-3.1-8B            & \texttt{meta-llama/Llama-3.1-8B} \\
Llama-3.1-8B-Instruct   & \texttt{meta-llama/Llama-3.1-8B-Instruct} \\
\midrule
\multicolumn{2}{l}{\textbf{Gemma 2 Family}} \\
\midrule
gemma-2-2b              & \texttt{google/gemma-2-2b} \\
gemma-2-2b-it           & \texttt{google/gemma-2-2b-it} \\
gemma-2-9b              & \texttt{google/gemma-2-9b} \\
gemma-2-9b-it           & \texttt{google/gemma-2-9b-it} \\
\midrule
\multicolumn{2}{l}{\textbf{Encoder-Decoder (Architecture Comparison)}} \\
\midrule
Flan-T5-small           & \texttt{google/flan-t5-small} \\
Flan-T5-base            & \texttt{google/flan-t5-base} \\
Flan-T5-large           & \texttt{google/flan-t5-large} \\
Flan-T5-xl              & \texttt{google/flan-t5-xl} \\
\midrule
\multicolumn{2}{l}{\textbf{Encoder-only (Architecture Comparison)}} \\
\midrule
BERT-base               & \texttt{google-bert/bert-base-uncased} \\
BERT-large              & \texttt{google-bert/bert-large-uncased} \\
RoBERTa-base            & \texttt{FacebookAI/roberta-base} \\
RoBERTa-large           & \texttt{FacebookAI/roberta-large} \\
ALBERT-base-v2          & \texttt{albert/albert-base-v2} \\
ALBERT-xlarge-v2        & \texttt{albert/albert-xlarge-v2} \\
\bottomrule
\end{tabular}
}
\caption{\textbf{Model information used in our experiments.} We also provide the Hugging Face ID.}
\label{tab:detail_model_info}
\end{table*}

Detailed information on the model is in Table~\ref{tab:detail_model_info}, which contains the Hugging Face ID of each model.

\subsection{Dataset information}
\label{app:datasets}

We infer using test sets for all benchmarks.

\paragraph{LogiQA} is a logical reasoning dataset sourced from the Chinese Civil Service Examination. It contains 651 test samples, each requiring multi-step logical inference over a given context. Questions cover various reasoning types, including categorical reasoning, conditional reasoning, and disjunctive reasoning.

\paragraph{SciQ} is a science question answering dataset containing 1,000 crowdsourced multiple-choice questions spanning physics, chemistry, and biology. Each question is paired with a short supporting passage (averaging $\sim$80 words) that provides the necessary context to answer correctly.

\paragraph{RACE (Reading comprehension from examinations)} consists of English reading comprehension questions collected from Chinese middle and high school exams. We utilize both subsets:
\begin{itemize}[leftmargin=*,topsep=0pt,parsep=0pt,itemsep=1.5pt]
    \item \textbf{RACE-M:} Middle school level with 1,436 test samples and simpler passages ($\sim$250 words on average)
    \item \textbf{RACE-H:} High school level with 3,498 test samples and more complex texts ($\sim$350 words on average)
\end{itemize}
Questions span various types, including detail retrieval, inference, vocabulary, and main idea identification.

\subsection{Prompting templates and evaluation protocols}
\label{app:prompts_eval}

We employ distinct prompting strategies and evaluation protocols tailored to each model architecture—Decoder-only (Causal LM), Encoder-only (Masked LM), and Encoder-Decoder (Seq2Seq)—to ensure fair and optimal performance assessment.

\subsubsection{Prompting templates}
We evaluate two primary template orderings to test context robustness: \textbf{CQO} (Context $\rightarrow$ Question $\rightarrow$ Options) and \textbf{QOC} (Question $\rightarrow$ Options $\rightarrow$ Context).

\paragraph{Decoder-only models}
For standard causal language models (e.g., Llama-3, Qwen, Gemma, Mistral), we use an open-ended completion format that guides the model to predict the next token as the answer label.

\begin{itemize}
    \item \textbf{CQO template}
    \begin{tcolorbox}[boxsep=2pt,colback=black!5,left=2pt,right=2pt,top=2pt,bottom=2pt]
    \textbf{\textit{Prompt}}: \\
    Context: \{Context\} \\
    Question: \{Question\} \\
    Options: \\
    A: \{Option A\} \\
    B: \{Option B\} \\
    C: \{Option C\} \\
    D: \{Option D\} \\
    Among A to D, the answer is:
    \end{tcolorbox}

    \item \textbf{QOC template}
    \begin{tcolorbox}[boxsep=2pt,colback=black!5,left=2pt,right=2pt,top=2pt,bottom=2pt]
    \textbf{\textit{Prompt}}: \\
    Question: \{Question\} \\
    Options: \\
    A: \{Option A\} \\
    B: \{Option B\} \\
    C: \{Option C\} \\
    D: \{Option D\} \\
    Context: \{Context\} \\
    Among A to D, the answer is:
    \end{tcolorbox}
\end{itemize}

\paragraph{Encoder-only models}
For masked language models (e.g., BERT, RoBERTa), we utilize a Cloze-style task where the model must predict the token at the \texttt{[MASK]} position.

\begin{itemize}
    \item \textbf{CQO template}
    \begin{tcolorbox}[boxsep=2pt,colback=black!5,left=2pt,right=2pt,top=2pt,bottom=2pt]
    \textbf{\textit{Prompt}}: \\
    Context: \{Context\} \\ \\
    Question: \{Question\} \\ \\
    Options: \\
    A. \{Option A\} \\
    B. \{Option B\} \\
    C. \{Option C\} \\
    D. \{Option D\} \\ \\
    Among A, B, C, D, the answer is \texttt{[MASK]}.
    \end{tcolorbox}

    \item \textbf{QOC template}
    \begin{tcolorbox}[boxsep=2pt,colback=black!5,left=2pt,right=2pt,top=2pt,bottom=2pt]
    \textbf{\textit{Prompt}}: \\
    Question: \{Question\} \\ \\
    Options: \\
    A. \{Option A\} \\
    B. \{Option B\} \\
    C. \{Option C\} \\
    D. \{Option D\} \\ \\
    Context: \{Context\} \\ \\
    Among A, B, C, D, the answer is \texttt{[MASK]}.
    \end{tcolorbox}
\end{itemize}

\paragraph{Encoder-Decoder models}
For sequence-to-sequence models (e.g., FLAN-T5), the full prompt is provided to the encoder. The decoder is initialized to generate the answer. We use the "Full" variant where the entire content resides in the encoder.

\begin{itemize}
    \item \textbf{CQO template}
    \begin{tcolorbox}[boxsep=2pt,colback=black!5,left=2pt,right=2pt,top=2pt,bottom=2pt]
    \textbf{\textit{Encoder Input}}: \\
    Context: \{Context\} \\ \\
    Question: \{Question\} \\ \\
    Options: \\
    A. \{Option A\} \\
    B. \{Option B\} \\
    C. \{Option C\} \\
    D. \{Option D\} \\ \\
    The answer is
    \end{tcolorbox}

    \item \textbf{QOC template}
    \begin{tcolorbox}[boxsep=2pt,colback=black!5,left=2pt,right=2pt,top=2pt,bottom=2pt]
    \textbf{\textit{Encoder Input}}: \\
    Question: \{Question\} \\ \\
    Options: \\
    A. \{Option A\} \\
    B. \{Option B\} \\
    C. \{Option C\} \\
    D. \{Option D\} \\ \\
    Context: \{Context\} \\ \\
    The answer is
    \end{tcolorbox}
\end{itemize}

\subsubsection{Evaluation protocols}
\label{sec:eval_protocols}

\noindent\textbf{Decoder-only models (likelihood scoring).} 
Instead of relying on unconstrained text generation, which requires complex parsing and may yield invalid outputs, we evaluate models using a constrained likelihood approach. 
Given the input prompt, we compute the logits for the next token prediction. We extract the logits corresponding to the valid option tokens (`A', `B', `C', `D') and their token variations. 
We apply a Softmax operation over these four values to obtain a normalized probability distribution:
\[ P(Correct) = \frac{e^{logit_{answer}}}{\sum_{k \in \{A,B,C,D\}} e^{logit_k}} \]
The model's prediction is the option with the highest probability. This method allows for a precise measurement of the model's preference even when differences are subtle.

\noindent\textbf{Encoder-only models (masked prediction).} 
We formulate the task as Masked Language Modeling (MLM). The model processes the bidirectional context and predicts the token at the \texttt{[MASK]} position. 
Similar to the decoder approach, we extract the probabilities for the tokens corresponding to the options `A', `B', `C', and `D'. 
To handle tokenizer differences, we select the maximum logit across case variations (e.g., 'A' and 'a') for each option. These maximum logits are then normalized via Softmax to obtain the final probability distribution.

\noindent\textbf{Encoder-Decoder models.} 
For T5-based models, we employ the same evaluation protocol as decoder-only models.



\subsection{Inference details}
\label{app:inference}

All experiments are conducted with greedy decoding single runs, maximum 16 output tokens, bfloat16 precision, and NVIDIA A6000 GPUs.

\section{Additional result}
\label{app:additional_results}

This section provides detailed per-model and per-dataset results for all experiments discussed in the main paper.

\subsection{Full model-dataset results}
\label{app:full_results}

~\Cref{tab:full_results} shows CQO and QOC accuracy for all 21 decoder-only models across 4 datasets. The average gap is +14.7\%.

\begin{table*}[t]
\centering

\resizebox{0.85\textwidth}{!}{%
\begin{tabular}{l|cc|cc|cc|cc|c}
\toprule
\multirow{2}{*}{\textbf{Model}} & \multicolumn{2}{c|}{\textbf{LogiQA}} & \multicolumn{2}{c|}{\textbf{SciQ}} & \multicolumn{2}{c|}{\textbf{RACE-M}} & \multicolumn{2}{c|}{\textbf{RACE-H}} & \textbf{Avg} \\
& CQO & QOC & CQO & QOC & CQO & QOC & CQO & QOC & Gap \\
\midrule
\multicolumn{10}{l}{\textit{LLaMA Family}} \\ 
Llama-3.1-8B & 39.8 & 33.8 & 96.3 & 91.8 & 77.6 & 47.1 & 73.9 & 48.2 & \textbf{+16.7} \\
Llama-3.1-8B-Instruct & 42.1 & 36.6 & 97.8 & 97.1 & 86.1 & 64.9 & 82.1 & 61.0 & \textbf{+12.1} \\
Llama-3.2-1B & 29.6 & 27.0 & 75.8 & 56.0 & 38.4 & 29.5 & 36.0 & 27.2 & \textbf{+10.0} \\
Llama-3.2-1B-Instruct & 31.0 & 29.6 & 91.9 & 83.5 & 60.7 & 43.2 & 57.3 & 41.2 & \textbf{+10.8} \\
Llama-3.2-3B & 27.3 & 23.7 & 93.4 & 84.3 & 64.0 & 39.3 & 60.0 & 37.6 & \textbf{+15.0} \\
Llama-3.2-3B-Instruct & 34.9 & 32.6 & 96.6 & 94.0 & 80.6 & 54.2 & 75.6 & 52.2 & \textbf{+13.7} \\
\midrule
\multicolumn{10}{l}{\textit{Qwen Family}} \\ 
Qwen2.5-0.5B & 28.7 & 29.2 & 83.6 & 69.0 & 58.4 & 40.0 & 51.4 & 41.3 & \textbf{+10.7} \\
Qwen2.5-0.5B-Instruct & 27.0 & 27.5 & 93.6 & 78.9 & 58.6 & 38.6 & 52.8 & 39.5 & \textbf{+11.9} \\
Qwen2.5-1.5B & 40.4 & 33.9 & 97.7 & 91.9 & 81.5 & 52.4 & 76.3 & 53.4 & \textbf{+16.1} \\
Qwen2.5-1.5B-Instruct & 42.2 & 34.3 & 97.0 & 90.6 & 80.8 & 47.8 & 75.8 & 48.7 & \textbf{+18.6} \\
Qwen2.5-3B & 42.5 & 33.0 & 97.7 & 93.2 & 87.3 & 55.9 & 82.4 & 57.0 & \textbf{+17.7} \\
Qwen2.5-3B-Instruct & 42.9 & 34.4 & 97.8 & 93.8 & 86.8 & 57.6 & 82.2 & 56.8 & \textbf{+16.8} \\
Qwen2.5-7B & 51.0 & 38.2 & 98.3 & 96.2 & 90.6 & 62.8 & 87.7 & 64.7 & \textbf{+16.4} \\
Qwen2.5-7B-Instruct & 53.1 & 39.8 & 98.6 & 97.4 & 90.1 & 68.1 & 87.3 & 65.2 & \textbf{+14.7} \\
Qwen3-0.6B & 37.2 & 26.4 & 87.4 & 69.8 & 51.2 & 33.9 & 46.9 & 32.2 & \textbf{+15.1} \\
Qwen3-1.7B & 41.0 & 36.7 & 94.7 & 86.3 & 74.9 & 44.4 & 70.9 & 45.6 & \textbf{+17.1} \\
Qwen3-4B & 52.1 & 39.3 & 98.1 & 95.5 & 86.8 & 55.4 & 80.8 & 55.2 & \textbf{+18.1} \\
\midrule
\multicolumn{10}{l}{\textit{Gemma Family}} \\ 
gemma-2-2b & 29.0 & 28.4 & 88.1 & 73.3 & 55.4 & 32.1 & 45.5 & 30.6 & \textbf{+13.4} \\
gemma-2-2b-it & 38.9 & 33.9 & 96.5 & 90.3 & 75.4 & 47.6 & 67.0 & 45.1 & \textbf{+15.2} \\
gemma-2-9b & 39.2 & 35.8 & 98.0 & 94.9 & 84.8 & 52.4 & 80.2 & 54.1 & \textbf{+16.3} \\
gemma-2-9b-it & 50.7 & 38.1 & 98.4 & 97.0 & 90.9 & 73.8 & 86.8 & 66.9 & \textbf{+12.7} \\
\midrule
\textbf{Average} & 39.1 & 33.0 & 94.1 & 86.9 & 74.3 & 49.6 & 69.5 & 48.8 & \textbf{+14.7} \\
\bottomrule
\end{tabular}
}
\caption{\textbf{Full accuracy results.} Performance of 21 decoder-only models on 4 datasets. Gap = CQO $-$ QOC.}
\label{tab:full_results}
\end{table*}

\subsection{Base vs instruction-tuned models}
\label{app:base_instruct}

~\Cref{tab:base_instruct_per_model} compares CQO-QOC gaps between base and instruction-tuned model pairs. Both variants show consistent gaps (Base: 14.70\%, Instruct: 14.12\%), indicating that instruction tuning does not mitigate order sensitivity.

\begin{table*}[t]
\centering
\resizebox{0.5\textwidth}{!}{%
\begin{tabular}{ll|ccc}
\toprule
\textbf{Model Pair} & \textbf{Dataset} & \textbf{Base Gap} & \textbf{Inst Gap} & \textbf{Diff} \\
\midrule

\multicolumn{5}{l}{\textit{LLaMA Family}} \\
\multirow{5}{*}{Llama-3.1-8B} & LogiQA  & +5.99  & +5.53  & -0.46 \\
 & SciQ   & +4.50  & +0.70  & -3.80 \\
 & RACE-M & +30.57 & +21.17 & -9.40 \\
 & RACE-H & +25.64 & +21.10 & -4.55 \\
 & \textbf{Average} & \textbf{+16.68} & \textbf{+12.12} & \textbf{-4.55} \\
\midrule

\multirow{5}{*}{Llama-3.2-1B} & LogiQA  & +2.61  & +1.38  & -1.23 \\
 & SciQ   & +19.80 & +8.40  & -11.40 \\
 & RACE-M & +8.91  & +17.48 & +8.57 \\
 & RACE-H & +8.75  & +16.09 & +7.35 \\
 & \textbf{Average} & \textbf{+10.02} & \textbf{+10.84} & \textbf{+0.82} \\
\midrule

\multirow{5}{*}{Llama-3.2-3B} & LogiQA  & +3.69  & +2.30  & -1.38 \\
 & SciQ   & +9.10  & +2.60  & -6.50 \\
 & RACE-M & +24.65 & +26.39 & +1.74 \\
 & RACE-H & +22.41 & +23.41 & +1.00 \\
 & \textbf{Average} & \textbf{+14.96} & \textbf{+13.68} & \textbf{-1.29} \\
\midrule

\multicolumn{5}{l}{\textit{Qwen Family}} \\
\multirow{5}{*}{Qwen2.5-0.5B} & LogiQA  & -0.46  & -0.46  & +0.00 \\
 & SciQ   & +14.60 & +14.70 & +0.10 \\
 & RACE-M & +18.31 & +19.99 & +1.67 \\
 & RACE-H & +10.15 & +13.32 & +3.17 \\
 & \textbf{Average} & \textbf{+10.65} & \textbf{+11.89} & \textbf{+1.24} \\
\midrule

\multirow{5}{*}{Qwen2.5-1.5B} & LogiQA  & +6.45  & +7.99  & +1.54 \\
 & SciQ   & +5.80  & +6.40  & +0.60 \\
 & RACE-M & +29.11 & +32.94 & +3.83 \\
 & RACE-H & +22.90 & +27.10 & +4.20 \\
 & \textbf{Average} & \textbf{+16.06} & \textbf{+18.61} & \textbf{+2.54} \\
\midrule

\multirow{5}{*}{Qwen2.5-3B} & LogiQA  & +9.52  & +10.29 & +0.77 \\
 & SciQ   & +4.50  & +4.20  & -0.30 \\
 & RACE-M & +31.96 & +29.25 & -2.72 \\
 & RACE-H & +25.24 & +25.36 & +0.11 \\
 & \textbf{Average} & \textbf{+17.81} & \textbf{+17.27} & \textbf{-0.53} \\
\midrule

\multirow{5}{*}{Qwen2.5-7B} & LogiQA  & +12.75 & +13.36 & +0.61 \\
 & SciQ   & +2.10  & +1.20  & -0.90 \\
 & RACE-M & +27.79 & +22.01 & -5.78 \\
 & RACE-H & +23.04 & +22.10 & -0.94 \\
 & \textbf{Average} & \textbf{+16.42} & \textbf{+14.67} & \textbf{-1.75} \\
\midrule

\multicolumn{5}{l}{\textit{Gemma Family}} \\
\multirow{5}{*}{Gemma-2-2B} & LogiQA  & +0.61  & +4.92  & +4.30 \\
 & SciQ   & +14.80 & +6.20  & -8.60 \\
 & RACE-M & +23.26 & +27.79 & +4.53 \\
 & RACE-H & +14.89 & +21.98 & +7.09 \\
 & \textbf{Average} & \textbf{+13.39} & \textbf{+15.22} & \textbf{+1.83} \\
\midrule

\multirow{5}{*}{Gemma-2-9B} & LogiQA  & +3.38  & +12.60 & +9.22 \\
 & SciQ   & +3.10  & +1.40  & -1.70 \\
 & RACE-M & +32.45 & +17.06 & -15.39 \\
 & RACE-H & +26.13 & +19.93 & -6.20 \\
 & \textbf{Average} & \textbf{+16.26} & \textbf{+12.75} & \textbf{-3.52} \\
\midrule

\textbf{Grand Average} & -- & \textbf{+14.70} & \textbf{+14.12} & \textbf{-0.58} \\
\bottomrule
\end{tabular}%
}
\caption{\textbf{CQO-QOC gap comparison: Base vs Instruction-tuned models (breakdown by dataset).} Both Base and Instruct models show similar ordering sensitivity. Diff = Instruct Gap $-$ Base Gap.}
\label{tab:base_instruct_per_model}
\end{table*}

\subsection{In-context learning results}
\label{app:icl}

~\Cref{tab:icl_per_model} shows QOC performance with 0, 1, 3, and 5 in-context examples. ICL provides marginal improvement (+3.1\% from 0-shot to 5-shot) but cannot close the gap to CQO (69.26\%).

\begin{table*}[t]
\centering
\resizebox{0.7\textwidth}{!}{%
\begin{tabular}{l|cccc|c}
\toprule
\textbf{Model} & \textbf{0-shot} & \textbf{1-shot} & \textbf{3-shot} & \textbf{5-shot} & \textbf{$\Delta$} \\
\midrule
\multicolumn{6}{l}{\textit{LLaMA Family}} \\ 
Llama-3.1-8B & 55.22 & 60.35 & 62.29 & 62.78 & +7.56 \\
Llama-3.1-8B-Instruct & 64.90 & 64.30 & 65.67 & 64.80 & -0.10 \\
Llama-3.2-1B & 34.93 & 28.05 & 31.93 & 32.64 & -2.30 \\
Llama-3.2-1B-Instruct & 49.39 & 46.59 & 50.68 & 51.54 & +2.15 \\
Llama-3.2-3B & 46.22 & 50.18 & 53.24 & 53.27 & +7.04 \\
Llama-3.2-3B-Instruct & 58.24 & 57.23 & 58.78 & 59.60 & +1.35 \\
\multicolumn{6}{l}{\textit{Qwen Family}} \\ 
Qwen2.5-0.5B & 44.87 & 44.29 & 45.06 & 45.37 & +0.50 \\
Qwen2.5-0.5B-Instruct & 46.13 & 41.46 & 40.30 & 42.45 & -3.68 \\
Qwen2.5-1.5B & 57.90 & 57.60 & 58.45 & 58.55 & +0.65 \\
Qwen2.5-1.5B-Instruct & 55.36 & 54.70 & 57.26 & 58.14 & +2.78 \\
Qwen2.5-3B & 59.75 & 59.49 & 61.26 & 61.06 & +1.31 \\
Qwen2.5-3B-Instruct & 60.49 & 59.22 & 61.36 & 61.92 & +1.42 \\
Qwen2.5-7B & 65.49 & 65.95 & 66.52 & 65.82 & +0.33 \\
Qwen2.5-7B-Instruct & 67.62 & 67.19 & 67.71 & 68.50 & +0.88 \\
Qwen3-0.6B & 40.60 & 44.08 & 50.17 & 51.65 & +11.05 \\
Qwen3-1.7B & 53.25 & 55.13 & 58.51 & 58.79 & +5.54 \\
Qwen3-4B & 61.36 & 64.86 & 67.25 & 67.15 & +5.79 \\
\multicolumn{6}{l}{\textit{Gemma Family}} \\ 
Gemma-2-2B & 41.10 & 47.84 & 49.94 & 49.89 & +8.80 \\
Gemma-2-2B-Instruct & 54.23 & 59.89 & 61.01 & 61.26 & +7.03 \\
Gemma-2-9B & 59.29 & 62.33 & 64.46 & 65.28 & +5.99 \\
Gemma-2-9B-Instruct & 68.95 & 67.72 & 70.09 & 69.75 & +0.79 \\
\midrule
\textbf{Average} & 54.54 & 55.16 & 57.24 & 57.63 & \textbf{+3.09} \\
\bottomrule
\end{tabular}}
\caption{\textbf{In-context learning results per model (QOC format).} Accuracy with 0, 1, 3, and 5 in-context examples. $\Delta$ = 5-shot $-$ 0-shot. ICL provides marginal improvement (+3.1\%) but cannot close the gap to CQO baseline (69.3\%).}
\label{tab:icl_per_model}
\end{table*}

\subsection{Option recall accuracy}
\label{app:recall}

~\Cref{tab:recall_per_model} shows option recall accuracy by dataset. High recall rates in both CQO (93.5\%) and QOC (94.7\%) confirm that the QOC performance drop is not due to memory failure.

\begin{table*}[t]
\centering
\resizebox{0.7\textwidth}{!}{%
\begin{tabular}{l|cc|cc|cc|cc|cc}
\toprule
& \multicolumn{2}{c|}{\textbf{LogiQA}} & \multicolumn{2}{c|}{\textbf{SciQ}} & \multicolumn{2}{c|}{\textbf{RACE-M}} & \multicolumn{2}{c|}{\textbf{RACE-H}} & \multicolumn{2}{c}{\textbf{Avg}} \\
\textbf{Model} & CQO & QOC & CQO & QOC & CQO & QOC & CQO & QOC & CQO & QOC \\
\midrule
Llama-3.1-8B & 93.3 & 94.5 & 99.8 & 99.8 & 97.3 & 97.2 & 96.9 & 96.9 & 96.8 & 97.1 \\
Llama-3.1-8B-Instruct & 96.7 & 96.8 & 99.8 & 99.8 & 97.0 & 98.3 & 96.5 & 98.5 & 97.5 & 98.3 \\
Llama-3.2-1B & 89.9 & 85.3 & 97.8 & 97.6 & 87.2 & 88.1 & 87.4 & 88.2 & 90.6 & 89.8 \\
Llama-3.2-1B-Instruct & 89.7 & 91.2 & 98.3 & 99.1 & 92.0 & 97.6 & 91.9 & 97.7 & 93.0 & 96.4 \\
Llama-3.2-3B & 94.0 & 92.9 & 99.6 & 98.0 & 95.4 & 94.0 & 94.5 & 92.8 & 95.9 & 94.4 \\
Llama-3.2-3B-Instruct & 93.1 & 93.7 & 95.2 & 99.1 & 91.2 & 97.9 & 88.4 & 97.8 & 92.0 & 97.1 \\
Qwen2.5-0.5B & 89.6 & 90.0 & 99.2 & 98.4 & 96.5 & 96.3 & 96.8 & 96.1 & 95.5 & 95.2 \\
Qwen2.5-0.5B-Instruct & 85.9 & 86.1 & 94.9 & 96.0 & 92.5 & 95.7 & 92.9 & 95.1 & 91.6 & 93.2 \\
Qwen2.5-1.5B & 88.7 & 87.7 & 93.8 & 96.2 & 93.4 & 95.5 & 94.9 & 95.2 & 92.7 & 93.6 \\
Qwen2.5-1.5B-Instruct & 86.9 & 85.6 & 95.0 & 97.0 & 91.4 & 94.4 & 92.3 & 95.0 & 91.4 & 93.0 \\
Qwen2.5-3B & 88.9 & 89.0 & 99.2 & 99.2 & 95.9 & 96.8 & 95.8 & 96.5 & 95.0 & 95.4 \\
Qwen2.5-3B-Instruct & 88.7 & 89.3 & 99.1 & 99.2 & 96.5 & 97.4 & 96.2 & 96.9 & 95.1 & 95.7 \\
Qwen2.5-7B & 89.7 & 89.6 & 99.5 & 99.4 & 96.4 & 96.6 & 96.2 & 96.4 & 95.5 & 95.5 \\
Qwen2.5-7B-Instruct & 86.9 & 87.3 & 99.2 & 99.4 & 96.2 & 96.8 & 95.7 & 96.1 & 94.5 & 94.9 \\
Qwen3-0.6B & 87.7 & 89.6 & 96.5 & 97.0 & 77.3 & 91.5 & 84.2 & 93.3 & 86.4 & 92.8 \\
Qwen3-1.7B & 91.0 & 92.7 & 99.3 & 99.6 & 97.0 & 97.9 & 97.3 & 98.1 & 96.1 & 97.1 \\
Qwen3-4B & 88.9 & 89.2 & 99.7 & 99.6 & 97.5 & 97.7 & 97.2 & 97.4 & 95.8 & 96.0 \\
gemma-2-2b-it & 84.0 & 84.3 & 93.5 & 89.6 & 76.7 & 84.3 & 85.0 & 89.5 & 84.8 & 86.9 \\
gemma-2-9b-it & 94.8 & 95.1 & 98.8 & 99.2 & 97.1 & 98.0 & 96.4 & 97.8 & 96.8 & 97.5 \\
\midrule
\textbf{Average} & 89.9 & 90.0 & 97.8 & 98.1 & 92.9 & 95.4 & 93.5 & 95.5 & \textbf{93.5} & \textbf{94.7} \\
\bottomrule
\end{tabular}
}
\caption{\textbf{Option recall accuracy (\%) by prompt order.} High recall rates across both CQO and QOC templates confirm that memory is not the bottleneck for the QOC performance drop.}
\label{tab:recall_per_model}
\end{table*}

\subsection{Encoder and encoder-decoder model results}
\label{app:encoder_results}

~\Cref{tab:encoder_results} shows results for encoder-only models and encoder-decoder models.


\begin{table*}[t]
\centering

\begin{minipage}[t]{0.49\textwidth}
\centering
\small
\resizebox{\linewidth}{!}{%
\begin{tabular}{ll|ccc}
\toprule
\textbf{Model} & \textbf{Dataset} & \textbf{CQO} & \textbf{QOC} & \textbf{Gap} \\
\midrule
\multicolumn{5}{l}{\textit{Encoder Only Models}} \\

\multirow{5}{*}{BERT-Base} & LogiQA & 20.89 & 19.20 & +1.69 \\
 & SciQ & 24.60 & 25.60 & -1.00 \\
 & RACE-M & 21.10 & 21.38 & -0.28 \\
 & RACE-H & 18.95 & 17.92 & +1.03 \\
 & \textbf{Average} & \textbf{21.39} & \textbf{21.03} & \textbf{+0.36} \\
\midrule

\multirow{5}{*}{BERT-Large} & LogiQA & 20.74 & 19.82 & +0.92 \\
 & SciQ & 23.90 & 26.50 & -2.60 \\
 & RACE-M & 23.68 & 21.73 & +1.95 \\
 & RACE-H & 21.04 & 18.78 & +2.26 \\
 & \textbf{Average} & \textbf{22.34} & \textbf{21.71} & \textbf{+0.63} \\
\midrule

\multirow{5}{*}{RoBERTa-Base} & LogiQA & 20.74 & 20.74 & +0.00 \\
 & SciQ & 28.90 & 27.80 & +1.10 \\
 & RACE-M & 21.17 & 20.89 & +0.28 \\
 & RACE-H & 16.67 & 16.41 & +0.26 \\
 & \textbf{Average} & \textbf{21.87} & \textbf{21.46} & \textbf{+0.41} \\
\midrule

\multirow{5}{*}{RoBERTa-Large} & LogiQA & 21.97 & 25.19 & -3.23 \\
 & SciQ & 28.60 & 28.90 & -0.30 \\
 & RACE-M & 21.73 & 23.47 & -1.74 \\
 & RACE-H & 16.87 & 17.50 & -0.63 \\
 & \textbf{Average} & \textbf{22.29} & \textbf{23.76} & \textbf{-1.47} \\
\midrule

\multirow{5}{*}{ALBERT-Base} & LogiQA & 20.28 & 20.12 & +0.15 \\
 & SciQ & 25.60 & 25.90 & -0.30 \\
 & RACE-M & 20.68 & 20.82 & -0.14 \\
 & RACE-H & 17.21 & 17.21 & +0.00 \\
 & \textbf{Average} & \textbf{20.94} & \textbf{21.01} & \textbf{-0.07} \\
\midrule

\multirow{5}{*}{ALBERT-XLarge} & LogiQA & 19.97 & 20.12 & -0.15 \\
 & SciQ & 26.30 & 26.10 & +0.20 \\
 & RACE-M & 20.82 & 20.61 & +0.21 \\
 & RACE-H & 17.30 & 17.27 & +0.03 \\
 & \textbf{Average} & \textbf{21.10} & \textbf{21.03} & \textbf{+0.07} \\
\midrule

\textbf{Encoder Avg} & -- & -- & -- & \textbf{-0.02} \\
\bottomrule
\end{tabular}%
}
\end{minipage}
\hfill
\begin{minipage}[t]{0.49\textwidth}
\centering
\small
\resizebox{\linewidth}{!}{%
\begin{tabular}{ll|ccc}
\toprule
\textbf{Model} & \textbf{Dataset} & \textbf{CQO} & \textbf{QOC} & \textbf{Gap} \\
\midrule
\multicolumn{5}{l}{\textit{Encoder-Decoder Models}} \\

\multirow{5}{*}{Flan-T5-Small} & LogiQA & 27.80 & 29.95 & -2.15 \\
 & SciQ & 88.70 & 84.20 & +4.50 \\
 & RACE-M & 51.74 & 44.64 & +7.10 \\
 & RACE-H & 42.71 & 38.02 & +4.69 \\
 & \textbf{Average} & \textbf{52.74} & \textbf{49.20} & \textbf{+3.54} \\
\midrule

\multirow{5}{*}{Flan-T5-Base} & LogiQA & 32.10 & 30.41 & +1.69 \\
 & SciQ & 94.60 & 93.60 & +1.00 \\
 & RACE-M & 76.25 & 72.91 & +3.34 \\
 & RACE-H & 66.52 & 63.15 & +3.37 \\
 & \textbf{Average} & \textbf{67.37} & \textbf{65.02} & \textbf{+2.35} \\
\midrule

\multirow{5}{*}{Flan-T5-Large} & LogiQA & 36.41 & 34.10 & +2.30 \\
 & SciQ & 96.40 & 96.00 & +0.40 \\
 & RACE-M & 85.31 & 84.19 & +1.11 \\
 & RACE-H & 80.45 & 76.56 & +3.89 \\
 & \textbf{Average} & \textbf{74.64} & \textbf{72.71} & \textbf{+1.93} \\
\midrule

\multirow{5}{*}{Flan-T5-XL} & LogiQA & 35.33 & 33.33 & +2.00 \\
 & SciQ & 97.40 & 97.10 & +0.30 \\
 & RACE-M & 90.18 & 88.86 & +1.32 \\
 & RACE-H & 86.68 & 84.73 & +1.94 \\
 & \textbf{Average} & \textbf{77.40} & \textbf{76.01} & \textbf{+1.39} \\
\midrule

\textbf{Enc-Dec Avg} & -- & -- & -- & \textbf{+2.30} \\
\bottomrule
\end{tabular}%
}
\end{minipage}
\caption{\textbf{Encoder-only and encoder-decoder model results (breakdown by dataset).} Bidirectional attention in encoder models and cross-attention in encoder-decoder models eliminate ordering sensitivity. Gap = CQO $-$ QOC.}
\label{tab:encoder_results}
\end{table*}

\subsection{Gradient attribution}
\label{app:attribution}

\begin{figure}[t]
\centering
\includegraphics[width=\linewidth]{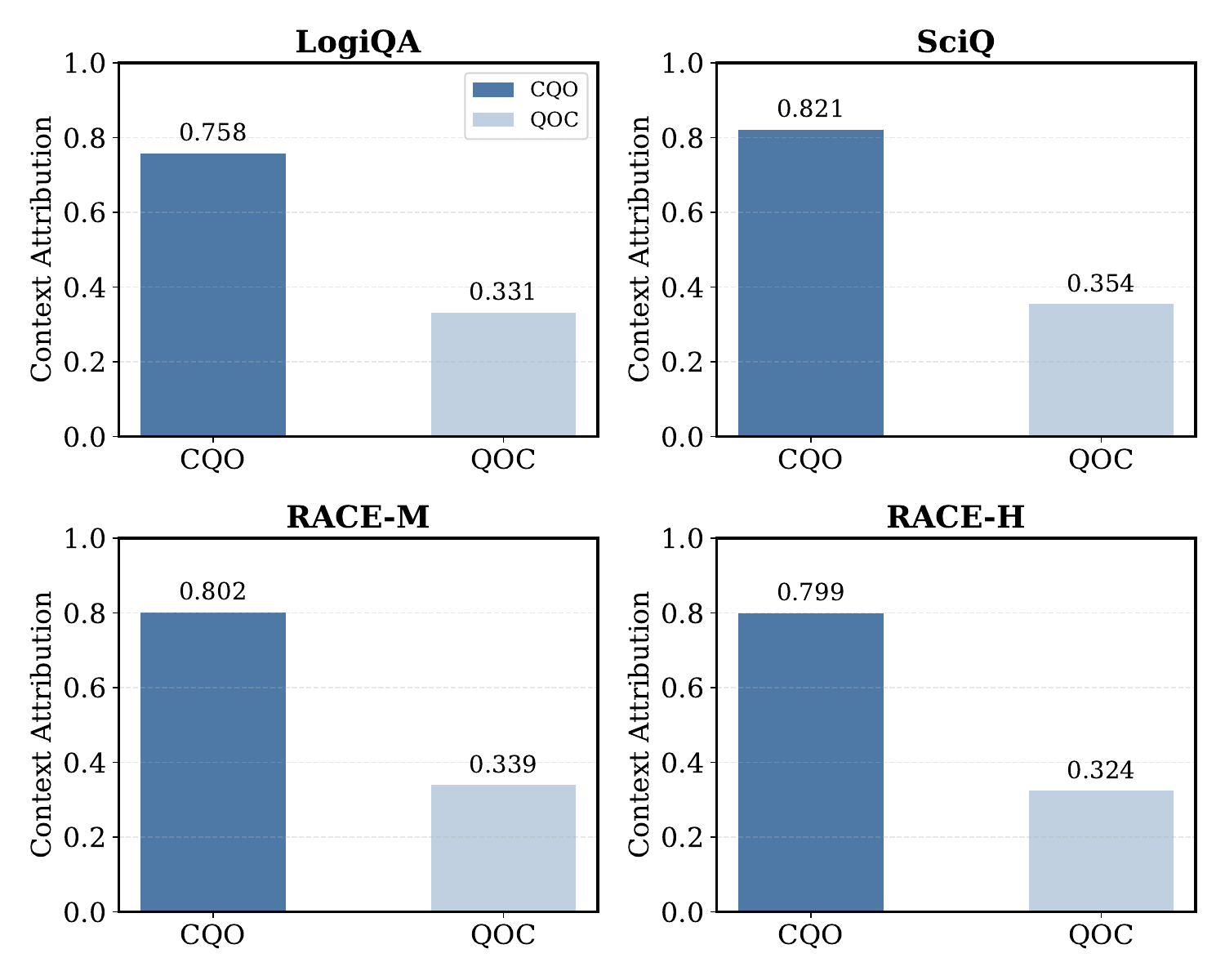}
\caption{\textbf{Context attribution ratio (CQO vs QOC) by dataset.} CQO consistently utilizes context more effectively across all benchmarks.}
\label{fig:attribution_by_dataset}
\end{figure}

~\Cref{fig:attribution_by_dataset} shows context attribution ratios by dataset. Across all datasets, CQO receives significantly more gradient flow from context tokens than QOC (average ratio: 2.38$\times$). We also show details in~\Cref{tab:attribution_per_model}.

\begin{table*}[t]
\centering

\begin{minipage}[t]{0.49\textwidth}
\centering
\small
\resizebox{\linewidth}{!}{%
\begin{tabular}{ll|ccc}
\toprule
\textbf{Model} & \textbf{Dataset} & \textbf{CQO} & \textbf{QOC} & \textbf{Ratio} \\
\midrule

\multicolumn{5}{l}{\textit{LLaMA Family}} \\
\multirow{5}{*}{Llama-3.1-8B} & LogiQA & 0.806 & 0.257 & 3.133$\times$ \\
 & SciQ & 0.769 & 0.269 & 2.859$\times$ \\
 & RACE-M & 0.859 & 0.368 & 2.335$\times$ \\
 & RACE-H & 0.881 & 0.400 & 2.205$\times$ \\
 & \textbf{Average} & \textbf{0.829} & \textbf{0.323} & \textbf{2.562$\times$} \\
\midrule

\multirow{5}{*}{Llama-3.1-8B-Instruct} & LogiQA & 0.835 & 0.248 & 3.364$\times$ \\
 & SciQ & 0.767 & 0.294 & 2.612$\times$ \\
 & RACE-M & 0.870 & 0.403 & 2.157$\times$ \\
 & RACE-H & 0.891 & 0.432 & 2.060$\times$ \\
 & \textbf{Average} & \textbf{0.841} & \textbf{0.344} & \textbf{2.441$\times$} \\
\midrule

\multirow{5}{*}{Llama-3.2-1B} & LogiQA & 0.797 & 0.255 & 3.129$\times$ \\
 & SciQ & 0.814 & 0.220 & 3.697$\times$ \\
 & RACE-M & 0.868 & 0.386 & 2.251$\times$ \\
 & RACE-H & 0.881 & 0.414 & 2.128$\times$ \\
 & \textbf{Average} & \textbf{0.840} & \textbf{0.319} & \textbf{2.637$\times$} \\
\midrule

\multirow{5}{*}{Llama-3.2-1B-Instruct} & LogiQA & 0.841 & 0.293 & 2.869$\times$ \\
 & SciQ & 0.808 & 0.299 & 2.700$\times$ \\
 & RACE-M & 0.889 & 0.413 & 2.154$\times$ \\
 & RACE-H & 0.906 & 0.424 & 2.137$\times$ \\
 & \textbf{Average} & \textbf{0.861} & \textbf{0.357} & \textbf{2.410$\times$} \\
\midrule

\multirow{5}{*}{Llama-3.2-3B} & LogiQA & 0.836 & 0.275 & 3.037$\times$ \\
 & SciQ & 0.804 & 0.267 & 3.007$\times$ \\
 & RACE-M & 0.868 & 0.405 & 2.142$\times$ \\
 & RACE-H & 0.893 & 0.429 & 2.080$\times$ \\
 & \textbf{Average} & \textbf{0.850} & \textbf{0.344} & \textbf{2.470$\times$} \\
\midrule

\multirow{5}{*}{Llama-3.2-3B-Instruct} & LogiQA & 0.877 & 0.298 & 2.942$\times$ \\
 & SciQ & 0.815 & 0.319 & 2.556$\times$ \\
 & RACE-M & 0.901 & 0.460 & 1.957$\times$ \\
 & RACE-H & 0.917 & 0.485 & 1.892$\times$ \\
 & \textbf{Average} & \textbf{0.877} & \textbf{0.390} & \textbf{2.247$\times$} \\
\midrule

\multicolumn{5}{l}{\textit{Qwen Family}} \\
\multirow{5}{*}{Qwen2.5-0.5B} & LogiQA & 0.855 & 0.306 & 2.798$\times$ \\
 & SciQ & 0.835 & 0.282 & 2.958$\times$ \\
 & RACE-M & 0.881 & 0.375 & 2.350$\times$ \\
 & RACE-H & 0.890 & 0.407 & 2.187$\times$ \\
 & \textbf{Average} & \textbf{0.865} & \textbf{0.342} & \textbf{2.526$\times$} \\
\midrule

\multirow{5}{*}{Qwen2.5-0.5B-Instruct} & LogiQA & 0.781 & 0.263 & 2.973$\times$ \\
 & SciQ & 0.816 & 0.250 & 3.260$\times$ \\
 & RACE-M & 0.843 & 0.399 & 2.111$\times$ \\
 & RACE-H & 0.850 & 0.428 & 1.987$\times$ \\
 & \textbf{Average} & \textbf{0.823} & \textbf{0.335} & \textbf{2.455$\times$} \\
\midrule

\multirow{5}{*}{Qwen2.5-1.5B} & LogiQA & 0.876 & 0.250 & 3.505$\times$ \\
 & SciQ & 0.823 & 0.219 & 3.752$\times$ \\
 & RACE-M & 0.884 & 0.312 & 2.834$\times$ \\
 & RACE-H & 0.892 & 0.325 & 2.745$\times$ \\
 & \textbf{Average} & \textbf{0.869} & \textbf{0.277} & \textbf{3.142$\times$} \\
\midrule

\multirow{5}{*}{Qwen2.5-1.5B-Instruct} & LogiQA & 0.825 & 0.210 & 3.926$\times$ \\
 & SciQ & 0.775 & 0.236 & 3.285$\times$ \\
 & RACE-M & 0.805 & 0.355 & 2.269$\times$ \\
 & RACE-H & 0.820 & 0.373 & 2.197$\times$ \\
 & \textbf{Average} & \textbf{0.806} & \textbf{0.293} & \textbf{2.747$\times$} \\
\bottomrule
\end{tabular}%
}
\end{minipage}
\hfill
\begin{minipage}[t]{0.49\textwidth}
\centering
\small
\resizebox{\linewidth}{!}{%
\begin{tabular}{ll|ccc}
\toprule
\textbf{Model} & \textbf{Dataset} & \textbf{CQO} & \textbf{QOC} & \textbf{Ratio} \\
\midrule

\multicolumn{5}{l}{\textit{Qwen Family (cont.)}} \\
\multirow{5}{*}{Qwen2.5-3B} & LogiQA & 0.871 & 0.224 & 3.886$\times$ \\
 & SciQ & 0.797 & 0.276 & 2.889$\times$ \\
 & RACE-M & 0.849 & 0.351 & 2.421$\times$ \\
 & RACE-H & 0.872 & 0.342 & 2.548$\times$ \\
 & \textbf{Average} & \textbf{0.848} & \textbf{0.298} & \textbf{2.841$\times$} \\
\midrule

\multirow{5}{*}{Qwen2.5-3B-Instruct} & LogiQA & 0.881 & 0.194 & 4.532$\times$ \\
 & SciQ & 0.832 & 0.281 & 2.965$\times$ \\
 & RACE-M & 0.873 & 0.374 & 2.333$\times$ \\
 & RACE-H & 0.890 & 0.387 & 2.298$\times$ \\
 & \textbf{Average} & \textbf{0.869} & \textbf{0.309} & \textbf{2.811$\times$} \\
\midrule

\multirow{5}{*}{Qwen2.5-7B} & LogiQA & 0.867 & 0.338 & 2.564$\times$ \\
 & SciQ & 0.868 & 0.368 & 2.360$\times$ \\
 & RACE-M & 0.890 & 0.479 & 1.859$\times$ \\
 & RACE-H & 0.907 & 0.500 & 1.814$\times$ \\
 & \textbf{Average} & \textbf{0.883} & \textbf{0.421} & \textbf{2.097$\times$} \\
\midrule

\multirow{5}{*}{Qwen2.5-7B-Instruct} & LogiQA & 0.857 & 0.256 & 3.350$\times$ \\
 & SciQ & 0.845 & 0.366 & 2.310$\times$ \\
 & RACE-M & 0.870 & 0.473 & 1.840$\times$ \\
 & RACE-H & 0.875 & 0.491 & 1.781$\times$ \\
 & \textbf{Average} & \textbf{0.862} & \textbf{0.396} & \textbf{2.174$\times$} \\
\midrule

\multirow{5}{*}{Qwen3-0.6B} & LogiQA & 0.860 & 0.189 & 4.553$\times$ \\
 & SciQ & 0.832 & 0.219 & 3.806$\times$ \\
 & RACE-M & 0.852 & 0.316 & 2.695$\times$ \\
 & RACE-H & 0.869 & 0.341 & 2.547$\times$ \\
 & \textbf{Average} & \textbf{0.853} & \textbf{0.266} & \textbf{3.206$\times$} \\
\midrule

\multirow{5}{*}{Qwen3-1.7B} & LogiQA & 0.849 & 0.197 & 4.301$\times$ \\
 & SciQ & 0.791 & 0.256 & 3.083$\times$ \\
 & RACE-M & 0.859 & 0.357 & 2.408$\times$ \\
 & RACE-H & 0.871 & 0.364 & 2.390$\times$ \\
 & \textbf{Average} & \textbf{0.842} & \textbf{0.294} & \textbf{2.868$\times$} \\
\midrule

\multirow{5}{*}{Qwen3-4B} & LogiQA & 0.928 & 0.179 & 5.178$\times$ \\
 & SciQ & 0.846 & 0.318 & 2.658$\times$ \\
 & RACE-M & 0.897 & 0.446 & 2.010$\times$ \\
 & RACE-H & 0.902 & 0.430 & 2.097$\times$ \\
 & \textbf{Average} & \textbf{0.893} & \textbf{0.343} & \textbf{2.601$\times$} \\
\midrule

\multicolumn{5}{l}{\textit{Gemma Family}} \\
\multirow{5}{*}{Gemma-2-2B} & LogiQA & 0.513 & 0.276 & 1.861$\times$ \\
 & SciQ & 0.480 & 0.233 & 2.061$\times$ \\
 & RACE-M & 0.589 & 0.361 & 1.633$\times$ \\
 & RACE-H & 0.604 & 0.396 & 1.524$\times$ \\
 & \textbf{Average} & \textbf{0.547} & \textbf{0.316} & \textbf{1.727$\times$} \\
\midrule

\multirow{5}{*}{Gemma-2-2B-Instruct} & LogiQA & 0.445 & 0.199 & 2.233$\times$ \\
 & SciQ & 0.462 & 0.294 & 1.570$\times$ \\
 & RACE-M & 0.582 & 0.406 & 1.433$\times$ \\
 & RACE-H & 0.593 & 0.441 & 1.345$\times$ \\
 & \textbf{Average} & \textbf{0.520} & \textbf{0.335} & \textbf{1.553$\times$} \\
\midrule

\multirow{5}{*}{Gemma-2-9B} & LogiQA & 0.519 & 0.247 & 2.095$\times$ \\
 & SciQ & 0.519 & 0.267 & 1.947$\times$ \\
 & RACE-M & 0.617 & 0.373 & 1.655$\times$ \\
 & RACE-H & 0.633 & 0.396 & 1.600$\times$ \\
 & \textbf{Average} & \textbf{0.572} & \textbf{0.321} & \textbf{1.784$\times$} \\
\midrule

\multirow{5}{*}{Gemma-2-9B-Instruct} & LogiQA & 0.524 & 0.262 & 2.000$\times$ \\
 & SciQ & 0.558 & 0.332 & 1.681$\times$ \\
 & RACE-M & 0.654 & 0.499 & 1.310$\times$ \\
 & RACE-H & 0.655 & 0.553 & 1.186$\times$ \\
 & \textbf{Average} & \textbf{0.598} & \textbf{0.411} & \textbf{1.453$\times$} \\
\midrule

\textbf{Grand Average} & -- & \textbf{0.797} & \textbf{0.335} & \textbf{2.380$\times$} \\
\bottomrule
\end{tabular}%
}
\end{minipage}
\caption{\textbf{Gradient attribution (context ratio) per model (breakdown by dataset).} CQO consistently allocates more gradient flow to context tokens than QOC. Ratio = CQO / QOC.}
\label{tab:attribution_per_model}
\end{table*}

\subsection{Intervention results}
\label{app:intervention_details}

\paragraph{Attention pruning (CQO Degradation).}
~\Cref{tab:zeroing_per_model} shows the effect of blocking option-to-context attention in CQO prompts. The average accuracy drop is -26.8\%, confirming this attention pathway is essential.


\begin{table*}[t]
\centering
\resizebox{0.98\textwidth}{!}{%
\begin{tabular}{l|ccc|ccc|ccc|ccc|c}
\toprule
\multirow{2}{*}{\textbf{Model}} &
\multicolumn{3}{c|}{\textbf{LogiQA}} &
\multicolumn{3}{c|}{\textbf{SciQ}} &
\multicolumn{3}{c|}{\textbf{RACE-M}} &
\multicolumn{3}{c|}{\textbf{RACE-H}} &
\textbf{Avg} \\
& CQO & Pruned & $\Delta$ & CQO & Pruned & $\Delta$ & CQO & Pruned & $\Delta$ & CQO & Pruned & $\Delta$ & $\Delta$ \\
\midrule

\multicolumn{14}{l}{\textit{LLaMA Family}} \\
Llama-3.1-8B &
39.78 & 30.88 & -8.91 &
96.30 & 89.90 & -6.40 &
77.65 & 51.95 & -25.70 &
73.87 & 49.60 & -24.27 &
\textbf{-16.32} \\
Llama-3.1-8B-Instruct &
42.09 & 35.79 & -6.30 &
97.80 & 93.70 & -4.10 &
86.07 & 58.98 & -27.09 &
82.13 & 58.75 & -23.38 &
\textbf{-15.22} \\
Llama-3.2-1B &
29.65 & 29.34 & -0.31 &
75.80 & 61.50 & -14.30 &
38.37 & 32.03 & -6.34 &
35.99 & 31.30 & -4.69 &
\textbf{-6.41} \\
Llama-3.2-1B-Instruct &
31.03 & 31.18 & +0.15 &
91.90 & 83.30 & -8.60 &
60.72 & 42.34 & -18.38 &
57.26 & 42.65 & -14.61 &
\textbf{-10.36} \\
Llama-3.2-3B &
27.34 & 30.41 & +3.07 &
93.40 & 76.70 & -16.70 &
64.00 & 42.69 & -21.31 &
60.01 & 41.80 & -18.21 &
\textbf{-13.29} \\
Llama-3.2-3B-Instruct &
34.87 & 33.64 & -1.23 &
96.60 & 90.00 & -6.60 &
80.57 & 53.83 & -26.74 &
75.64 & 53.72 & -21.93 &
\textbf{-14.12} \\
\midrule

\multicolumn{14}{l}{\textit{Qwen Family}} \\
Qwen2.5-0.5B &
28.73 & 27.34 & -1.38 &
83.60 & 25.60 & -58.00 &
58.36 & 26.60 & -31.75 &
51.40 & 25.96 & -25.44 &
\textbf{-29.15} \\
Qwen2.5-0.5B-Instruct &
27.04 & 26.27 & -0.77 &
93.60 & 26.30 & -67.30 &
58.64 & 25.77 & -32.87 &
52.80 & 25.07 & -27.73 &
\textbf{-32.17} \\
Qwen2.5-1.5B &
40.40 & 27.80 & -12.60 &
97.70 & 29.70 & -68.00 &
81.48 & 26.74 & -54.74 &
76.27 & 25.53 & -50.74 &
\textbf{-46.52} \\
Qwen2.5-1.5B-Instruct &
42.24 & 25.81 & -16.44 &
97.00 & 31.50 & -65.50 &
80.78 & 27.09 & -53.69 &
75.84 & 26.44 & -49.40 &
\textbf{-46.26} \\
Qwen2.5-3B &
42.55 & 27.34 & -15.21 &
97.80 & 24.50 & -73.30 &
87.47 & 26.32 & -61.14 &
82.42 & 24.87 & -57.55 &
\textbf{-51.80} \\
Qwen2.5-3B-Instruct &
44.09 & 28.57 & -15.51 &
97.80 & 25.60 & -72.20 &
86.91 & 26.67 & -60.24 &
82.28 & 25.90 & -56.38 &
\textbf{-51.08} \\
Qwen2.5-7B &
51.00 & 31.03 & -19.97 &
98.30 & 79.90 & -18.40 &
90.60 & 48.89 & -41.71 &
87.74 & 45.03 & -42.71 &
\textbf{-30.70} \\
Qwen2.5-7B-Instruct &
53.15 & 35.33 & -17.82 &
98.60 & 81.30 & -17.30 &
90.11 & 46.87 & -43.25 &
87.31 & 46.20 & -41.11 &
\textbf{-29.87} \\
Qwen3-0.6B &
37.17 & 21.97 & -15.21 &
87.40 & 27.30 & -60.10 &
51.18 & 21.59 & -29.60 &
46.88 & 21.47 & -25.41 &
\textbf{-32.58} \\
Qwen3-1.7B &
41.01 & 26.57 & -14.44 &
94.87 & 30.40 & -64.47 &
74.86 & 26.32 & -48.54 &
70.87 & 27.90 & -42.97 &
\textbf{-42.60} \\
Qwen3-4B &
52.07 & 26.42 & -25.65 &
98.10 & 43.10 & -55.00 &
86.84 & 26.39 & -60.45 &
80.85 & 28.67 & -52.17 &
\textbf{-48.32} \\
\midrule

\multicolumn{14}{l}{\textit{Gemma Family}} \\
Gemma-2-2B &
29.03 & 27.65 & -1.38 &
88.10 & 80.60 & -7.50 &
55.36 & 39.55 & -15.81 &
45.45 & 40.42 & -5.03 &
\textbf{-7.43} \\
Gemma-2-2B-Instruct &
38.86 & 37.79 & -1.08 &
96.50 & 89.90 & -6.60 &
75.42 & 52.37 & -23.05 &
67.04 & 53.06 & -13.98 &
\textbf{-11.18} \\
Gemma-2-9B &
39.17 & 35.64 & -3.53 &
98.00 & 94.40 & -3.60 &
84.82 & 58.70 & -26.11 &
80.25 & 57.86 & -22.38 &
\textbf{-13.91} \\
Gemma-2-9B-Instruct &
50.69 & 43.32 & -7.37 &
98.40 & 94.90 & -3.50 &
90.88 & 68.25 & -22.63 &
86.82 & 66.92 & -19.90 &
\textbf{-13.35} \\
\midrule

\textbf{Average} &
\textbf{39.14} & \textbf{30.48} & \textbf{-8.66} &
\textbf{94.17} & \textbf{60.96} & \textbf{-33.21} &
\textbf{74.34} & \textbf{39.52} & \textbf{-34.82} &
\textbf{69.48} & \textbf{39.01} & \textbf{-30.48} &
\textbf{-26.79} \\
\bottomrule
\end{tabular}%
}
\caption{\textbf{Attention pruning results per model.} Blocking option-to-context attention in CQO prompts causes significant accuracy drop, confirming context utilization is essential. $\Delta$ = Zeroed $-$ CQO.}
\label{tab:zeroing_per_model}
\end{table*}

\paragraph{Activation patching (QOC Improvement).}
~\Cref{tab:patching_per_model} shows the effect of replacing QOC option hidden states with CQO representations. Average improvement: +6.0\%.

\begin{table*}[t]
\centering
\resizebox{0.85\textwidth}{!}{%
\begin{tabular}{l|cc|cc|cc|cc|c}
\toprule
\multirow{2}{*}{\textbf{Model}} & \multicolumn{2}{c|}{\textbf{LogiQA}} & \multicolumn{2}{c|}{\textbf{SciQ}} & \multicolumn{2}{c|}{\textbf{RACE-M}} & \multicolumn{2}{c|}{\textbf{RACE-H}} & \textbf{Avg} \\
& QOC & Patched & QOC & Patched & QOC & Patched & QOC & Patched & $\Delta$ \\
\midrule
\multicolumn{10}{l}{\textit{LLaMA Family}} \\
Llama-3.1-8B & 33.79 & 30.57 & 91.80 & 87.20 & 47.08 & 52.58 & 48.23 & 48.77 & \textbf{-0.45} \\
Llama-3.1-8B-Instruct & 36.56 & 28.42 & 97.10 & 94.70 & 64.90 & 61.98 & 61.03 & 51.97 & \textbf{-5.63} \\
Llama-3.2-1B & 27.04 & 26.73 & 56.00 & 57.00 & 29.46 & 29.53 & 27.24 & 28.70 & \textbf{+0.56} \\
Llama-3.2-1B-Instruct & 29.65 & 30.11 & 83.50 & 82.40 & 43.25 & 47.08 & 41.17 & 43.77 & \textbf{+1.45} \\
Llama-3.2-3B & 23.66 & 30.72 & 84.30 & 73.20 & 39.35 & 41.09 & 37.59 & 36.45 & \textbf{-0.86} \\
Llama-3.2-3B-Instruct & 32.57 & 31.80 & 94.00 & 93.40 & 54.18 & 55.78 & 52.23 & 47.97 & \textbf{-1.01} \\
\midrule
\multicolumn{10}{l}{\textit{Qwen Family}} \\
Qwen2.5-0.5B & 29.19 & 26.42 & 69.00 & 76.60 & 40.04 & 41.71 & 41.25 & 37.14 & \textbf{+0.60} \\
Qwen2.5-0.5B-Instruct & 27.50 & 23.96 & 78.90 & 78.70 & 38.65 & 36.98 & 39.48 & 32.76 & \textbf{-3.03} \\
Qwen2.5-1.5B & 33.95 & 37.02 & 91.90 & 95.00 & 52.37 & 71.24 & 53.37 & 64.69 & \textbf{+9.09} \\
Qwen2.5-1.5B-Instruct & 34.25 & 37.48 & 90.60 & 92.10 & 47.84 & 67.34 & 48.74 & 61.61 & \textbf{+9.27} \\
Qwen2.5-3B & 33.03 & 34.72 & 93.30 & 93.30 & 55.50 & 70.96 & 57.18 & 66.98 & \textbf{+6.74} \\
Qwen2.5-3B-Instruct & 33.79 & 31.34 & 93.60 & 96.20 & 57.66 & 72.70 & 56.92 & 66.21 & \textbf{+6.12} \\
Qwen2.5-7B & 38.25 & 46.08 & 96.20 & 98.30 & 62.81 & 88.72 & 64.69 & 83.28 & \textbf{+13.61} \\
Qwen2.5-7B-Instruct & 39.78 & 52.38 & 97.40 & 98.10 & 68.11 & 87.67 & 65.21 & 83.13 & \textbf{+12.70} \\
Qwen3-0.6B & 26.42 & 31.95 & 69.80 & 83.20 & 33.91 & 56.27 & 32.25 & 50.71 & \textbf{+14.94} \\
Qwen3-1.7B & 36.71 & 38.86 & 86.30 & 92.00 & 44.43 & 71.80 & 45.57 & 68.21 & \textbf{+14.46} \\
Qwen3-4B & 39.32 & 50.23 & 95.50 & 97.90 & 55.43 & 84.96 & 55.17 & 78.73 & \textbf{+16.60} \\
\midrule
\multicolumn{10}{l}{\textit{Gemma Family}} \\
Gemma-2-2B & 28.42 & 28.26 & 73.30 & 89.10 & 32.10 & 53.06 & 30.56 & 45.28 & \textbf{+12.83} \\
Gemma-2-2B-Instruct & 33.95 & 38.56 & 90.30 & 94.50 & 47.63 & 72.70 & 45.05 & 67.64 & \textbf{+14.12} \\
Gemma-2-9B & 35.79 & 32.72 & 94.90 & 71.80 & 52.37 & 62.12 & 54.12 & 59.23 & \textbf{-2.83} \\
Gemma-2-9B-Instruct & 38.10 & 50.38 & 97.00 & 95.80 & 73.82 & 80.43 & 66.90 & 72.44 & \textbf{+5.81} \\
\midrule
\textbf{Average} & 33.42 & 34.76 & 89.62 & 90.96 & 49.72 & 65.69 & 48.78 & 61.48 & \textbf{+5.96} \\
\bottomrule
\end{tabular}%
}
\caption{\textbf{Activation patching results.} Replacing QOC option hidden states with CQO representations improves accuracy, demonstrating that context-aware representations enhance prediction. $\Delta$ = Patched $-$ QOC.}
\label{tab:patching_per_model}
\end{table*}

\paragraph{Option repetition (QOCO).}
~\Cref{tab:qoco_results} shows the effect of repeating options after context (Q-O-C-O format). This simple modification allows the repeated options to attend to context, improving QOC accuracy by an average of +8.2\%.

\begin{table*}[t]
\centering
\resizebox{0.98\textwidth}{!}{%
\begin{tabular}{l|ccc|ccc|ccc|ccc|c}
\toprule
\multirow{2}{*}{\textbf{Model}}
& \multicolumn{3}{c|}{\textbf{LogiQA}}
& \multicolumn{3}{c|}{\textbf{SciQ}}
& \multicolumn{3}{c|}{\textbf{RACE-M}}
& \multicolumn{3}{c|}{\textbf{RACE-H}}
& \textbf{Avg} \\
& QOC & QOCO & $\Delta$
& QOC & QOCO & $\Delta$
& QOC & QOCO & $\Delta$
& QOC & QOCO & $\Delta$
& $\Delta$ \\
\midrule

\multicolumn{14}{l}{\textit{LLaMA Family}} \\
Llama-3.1-8B & 33.79 & 38.40 & +4.61 & 91.80 & 86.90 & -4.90 & 47.08 & 59.19 & +12.11 & 48.23 & 58.98 & +10.75 & \textbf{+5.64} \\
Llama-3.1-8B-Instruct & 36.56 & 39.48 & +2.92 & 97.10 & 97.40 & +0.30 & 64.90 & 78.55 & +13.65 & 61.03 & 73.79 & +12.76 & \textbf{+7.40} \\
Llama-3.2-1B & 27.04 & 27.19 & +0.15 & 56.00 & 68.50 & +12.50 & 29.46 & 34.26 & +4.80 & 27.24 & 32.08 & +4.84 & \textbf{+5.57} \\
Llama-3.2-1B-Instruct & 29.65 & 29.03 & -0.62 & 83.50 & 94.20 & +10.70 & 43.25 & 55.85 & +12.60 & 41.17 & 51.43 & +10.26 & \textbf{+8.23} \\
Llama-3.2-3B & 23.66 & 30.57 & +6.91 & 84.30 & 92.40 & +8.10 & 39.35 & 53.76 & +14.41 & 37.59 & 47.66 & +10.07 & \textbf{+9.87} \\
Llama-3.2-3B-Instruct & 32.57 & 35.48 & +2.91 & 94.00 & 96.50 & +2.50 & 54.18 & 72.14 & +17.96 & 52.23 & 66.47 & +14.24 & \textbf{+9.40} \\
\midrule

\multicolumn{14}{l}{\textit{Qwen Family}} \\
Qwen2.5-0.5B & 29.19 & 27.50 & -1.69 & 69.00 & 85.10 & +16.10 & 40.04 & 43.31 & +3.27 & 41.25 & 38.74 & -2.51 & \textbf{+3.79} \\
Qwen2.5-0.5B-Instruct & 27.50 & 24.27 & -3.23 & 78.90 & 88.70 & +9.80 & 38.65 & 48.40 & +9.75 & 39.48 & 44.57 & +5.09 & \textbf{+5.35} \\
Qwen2.5-1.5B & 33.95 & 33.95 & 0.00 & 91.90 & 95.60 & +3.70 & 52.37 & 63.58 & +11.21 & 53.37 & 56.43 & +3.06 & \textbf{+4.49} \\
Qwen2.5-1.5B-Instruct & 34.25 & 35.02 & +0.77 & 90.60 & 91.40 & +0.80 & 47.84 & 60.38 & +12.54 & 48.74 & 59.75 & +11.01 & \textbf{+6.28} \\
Qwen2.5-3B & 33.03 & 35.94 & +2.91 & 93.30 & 97.10 & +3.80 & 55.50 & 71.31 & +15.81 & 57.18 & 66.18 & +9.00 & \textbf{+7.88} \\
Qwen2.5-3B-Instruct & 33.79 & 38.56 & +4.77 & 93.60 & 96.80 & +3.20 & 57.66 & 73.75 & +16.09 & 56.92 & 69.01 & +12.09 & \textbf{+9.04} \\
Qwen2.5-7B & 38.25 & 42.70 & +4.45 & 96.20 & 98.00 & +1.80 & 62.81 & 79.18 & +16.37 & 64.69 & 74.76 & +10.07 & \textbf{+8.17} \\
Qwen2.5-7B-Instruct & 39.78 & 45.78 & +6.00 & 97.40 & 97.50 & +0.10 & 68.11 & 80.99 & +12.88 & 65.21 & 76.90 & +11.69 & \textbf{+7.66} \\
Qwen3-0.6B & 26.42 & 34.56 & +8.14 & 69.80 & 80.10 & +10.30 & 33.91 & 49.86 & +15.95 & 32.25 & 45.20 & +12.95 & \textbf{+11.83} \\
Qwen3-1.7B & 36.71 & 36.10 & -0.61 & 86.30 & 82.10 & -4.20 & 44.43 & 55.99 & +11.56 & 45.57 & 56.98 & +11.41 & \textbf{+4.54} \\
Qwen3-4B & 39.32 & 48.54 & +9.22 & 95.50 & 98.20 & +2.70 & 55.43 & 76.74 & +21.31 & 55.17 & 72.41 & +17.24 & \textbf{+12.61} \\
\midrule

\multicolumn{14}{l}{\textit{Gemma Family}} \\
Gemma-2-2B & 28.42 & 26.11 & -2.31 & 73.30 & 91.60 & +18.30 & 32.10 & 48.05 & +15.95 & 30.56 & 45.25 & +14.69 & \textbf{+11.65} \\
Gemma-2-2B-Instruct & 33.95 & 41.01 & +7.06 & 90.30 & 94.90 & +4.60 & 47.63 & 71.80 & +24.17 & 45.05 & 65.84 & +20.79 & \textbf{+14.15} \\
Gemma-2-9B & 35.79 & 38.25 & +2.46 & 94.90 & 97.60 & +2.70 & 52.37 & 71.80 & +19.43 & 54.12 & 69.07 & +14.95 & \textbf{+9.88} \\
Gemma-2-9B-Instruct & 38.10 & 48.08 & +9.98 & 97.00 & 97.30 & +0.30 & 73.82 & 85.65 & +11.83 & 66.90 & 81.25 & +14.35 & \textbf{+9.11} \\
\midrule

\textbf{Average} & 32.93 & 36.02 & \textbf{+3.08} & 86.89 & 91.80 & \textbf{+4.91} & 49.56 & 63.54 & \textbf{+13.98} & 48.75 & 59.65 & \textbf{+10.89} & \textbf{+8.21} \\
\bottomrule
\end{tabular}%
}
\caption{\textbf{QOCO results per model (breakdown by dataset).} QOCO prompts (repeat variant) achieve improved accuracy to QOC, indicated by small $\Delta$. $\Delta$ = QOCO $-$ QOC.}
\label{tab:qoco_results}
\end{table*}

\section{Results on other architectures}
\label{app:other_arch}

To isolate \emph{causal masking} from \emph{softmax attention} as the source of the CQO--QOC gap, we extend our architecture comparison beyond Transformers to four additional model families: small-scale Mamba SSMs, FalconMamba-7B, the Qwen3-Next hybrid linear-attention model, and the diffusion-based Dream language model. All experiments use the same templates, datasets, and likelihood-scoring protocol as the main decoder-only evaluation. Results are summarized in~\Cref{tab:other_architectures}.

\begin{table}[h]
\centering
\small
\setlength{\tabcolsep}{3pt}
\resizebox{\linewidth}{!}{%
\begin{tabular}{llcccc}
\toprule
\textbf{Architecture} & \textbf{Model} & \textbf{CQO} & \textbf{QOC} & \textbf{QO} & \textbf{Gap} \\
\midrule
\multicolumn{6}{l}{\textit{State-space models (left-to-right, no softmax):}} \\
Mamba-small & avg. of 130M--2.8B (5 models) & 24.37 & 23.44 & 24.32 & $+0.93$ \\
Mamba-7B & FalconMamba-7B (base) & 72.19 & 53.96 & 56.75 & $+18.23$ \\
Mamba-7B & FalconMamba-7B (instruct) & 73.97 & 55.23 & 57.05 & $+18.74$ \\
\midrule
\multicolumn{6}{l}{\textit{Hybrid linear + softmax attention:}} \\
Qwen3-Next & Qwen3-Next-80B-A3B-Instruct (logicqa) & 66.82 & 46.70 & 47.00 & $+20.12$ \\
\midrule
\multicolumn{6}{l}{\textit{Diffusion LM with bidirectional-within-block masking:}} \\
Dream & Dream-v0-Base-7B & 77.56 & 77.99 & 61.64 & $-0.43$ \\
Dream & Dream-v0-Instruct-7B & 79.60 & 78.10 & 61.47 & $+1.50$ \\
\bottomrule
\end{tabular}
}
\caption{\textbf{Left-to-right non-Transformer architectures reproduce the CQO--QOC gap; only bidirectional-within-block Dream does not.} Small Mamba models are near-random on MCQA and therefore uninformative; the capable 7B FalconMamba and 80B Qwen3-Next---both still strictly autoregressive---show gaps on par with or larger than decoder-only Transformers, while Dream, whose within-block mask is bidirectional, shows essentially zero gap. The pattern pinpoints strictly left-to-right token processing, not softmax attention specifically, as the source of the bottleneck.}
\label{tab:other_architectures}
\end{table}

\paragraph{Takeaways.}
(i) Small-scale Mamba models are near-random on MCQA and thus cannot be used to decide the question, matching the encoder-only control pattern.
(ii) FalconMamba-7B, which solves the task competently, shows an \emph{even larger} gap than decoder-only Transformers ($+18.5\%$ vs.\ $+14.7\%$), strong evidence that sequential, left-to-right state compression---not softmax attention specifically---is what blocks the context from reaching option representations.
(iii) Qwen3-Next, which mixes gated linear attention with softmax attention in a hybrid block design but remains left-to-right, shows the largest gap we observe ($+20.1\%$ on LogiQA).
(iv) Only Dream, which uses bidirectional-within-block masking during its diffusion-style denoising, shows a near-zero gap despite being a 7B decoder-style model. Taken together, these results corroborate the causal-mask hypothesis and generalize it: any architecture that enforces strictly preceding-token access exhibits the CQO--QOC gap, while bidirectional (even within-block) access removes it.

\section{Closed-source model: Gemini}
\label{app:closed_source}

To test whether the CQO--QOC gap persists at commercial scale and whether chain-of-thought (CoT) prompting generalizes to strong closed-source models, we evaluate \textsc{gemini-3-flash-preview} (no-thinking mode) through the Vertex AI API. We use the same four MCQA benchmarks as the main paper and parse the model's A/B/C/D choice from the generated response via regex. Results are reported in~\Cref{tab:gemini_results}.

\begin{table}[h]
\centering
\small
\setlength{\tabcolsep}{4pt}
\resizebox{\linewidth}{!}{%
\begin{tabular}{llcccc}
\toprule
\textbf{Mode} & \textbf{Dataset} & \textbf{CQO} & \textbf{QOC} & \textbf{QO} & \textbf{Gap} \\
\midrule
\multirow{5}{*}{Baseline}
& LogiQA      & 77.88 & 66.21 & 53.00 & $+11.67$ \\
& SciQ        & 98.80 & 96.80 & 92.70 & $+2.00$  \\
& RACE-M      & 89.48 & 73.89 & 58.22 & $+15.60$ \\
& RACE-H      & 83.22 & 65.12 & 61.81 & $+18.10$ \\
& \textbf{Average} & \textbf{87.35} & \textbf{75.50} & \textbf{66.43} & $\mathbf{+11.84}$ \\
\midrule
\multirow{5}{*}{CoT}
& LogiQA      & 81.87 & 78.03 & 53.30 & $+3.84$ \\
& SciQ        & 99.30 & 99.20 & 97.50 & $+0.10$ \\
& RACE-M      & 96.38 & 94.92 & 68.59 & $+1.46$ \\
& RACE-H      & 94.05 & 91.57 & 72.16 & $+2.49$ \\
& \textbf{Average} & \textbf{92.90} & \textbf{90.93} & \textbf{72.89} & $\mathbf{+1.97}$ \\
\bottomrule
\end{tabular}
}
\caption{\textbf{Gemini exhibits the same CQO--QOC gap, and CoT prompting reduces it by 83.4\%.} Baseline gap: $+11.84$; CoT gap: $+1.97$. Shift-0 evaluation only. Consistent with the open-weight decoder-only models, CQO remains higher than QOC on every dataset, confirming that the effect is not an artifact of any particular model family or training recipe. The larger relative reduction under CoT also suggests that stronger models can better exploit the additional decoding budget unlocked by step-by-step reasoning.}
\label{tab:gemini_results}
\end{table}

\paragraph{Takeaways.}
Gemini, a production-grade commercial model, reproduces the ordering gap ($+11.84\%$) despite a much larger base accuracy (CQO $87.4\%$, QOC $75.5\%$), ruling out scale or training-data differences as the explanation. CoT prompting closes the gap by $83.4\%$ ($+11.84 \to +1.97$), a larger relative reduction than we observe for open-weight models ($49.2\%$). This is consistent with the hypothesis that stronger reasoning capability better exploits the extra decoding steps that CoT grants---the very mechanism predicted by the single-step-bottleneck account in Appendix~\ref{app:theory}.

\section{Formal analysis of the single-step bottleneck}
\label{app:theory}

This appendix gives a compact information-theoretic statement of the claim made in the main text: under the QOC ordering, the option representations are \emph{structurally} independent of the context, so evidence--option matching must be deferred to the final decoding step. The argument is architecture-agnostic and applies to any strictly autoregressive (left-to-right) sequence model.

\paragraph{Setup.}
Consider a decoder-only language model that maps an input sequence $x_{1:T}$ to a sequence of hidden states $h_{1:T} \in \mathbb{R}^{T \times d}$ at the final layer. Causal masking imposes the constraint
\begin{equation}
    h_t = f_\theta(x_{1:t}), \qquad t = 1, \dots, T,
\label{eq:causal_constraint}
\end{equation}
i.e.\ $h_t$ is a function of the prefix $x_{1:t}$ only. We denote by $C$, $Q$, $O = (O_1, \dots, O_K)$, and $A$ the context, question, options, and final answer-scoring position, respectively, and we write $h_O$ for the stack of hidden states at the option token positions.

\paragraph{Claim 1 (QOC: option representations are context-blind).}
Under the QOC template $x = [Q, O, C, A]$, equation~\eqref{eq:causal_constraint} implies $h_O = f_\theta(Q, O)$, and in particular
\begin{equation}
    I\!\left( h_O^{\mathrm{QOC}} ; \, C \;\big|\; Q, O \right) \;=\; 0,
\label{eq:mi_qoc}
\end{equation}
because $h_O^{\mathrm{QOC}}$ is a deterministic function of $(Q, O)$ alone. In words, no information about the context is representable in the option hidden states.

\paragraph{Claim 2 (CQO: option representations are context-conditioned).}
Under the CQO template $x = [C, Q, O, A]$, equation~\eqref{eq:causal_constraint} instead yields $h_O = f_\theta(C, Q, O)$, so there is no structural bound on $I( h_O^{\mathrm{CQO}} ; C \mid Q, O)$. Empirically, gradient$\times$input attribution assigns $0.797$ of total attribution to context tokens under CQO versus $0.335$ under QOC (\Cref{fig:fig6-b}), and attention to option tokens evolves in the opposite direction across depth in the two orderings (\Cref{fig:fig6-a}), consistent with this qualitative difference.

\paragraph{Consequence (single-step bottleneck).}
All evidence--option comparison under QOC is therefore performed through the \emph{final} answer-scoring step alone: $P(A \mid x) = g_\theta(h_A^{\mathrm{QOC}})$ with $h_A^{\mathrm{QOC}} = f_\theta(Q, O, C)$. Although $h_A$ does attend to both $O$ and $C$, the option representations it summarizes are context-blind by~\eqref{eq:mi_qoc}, so the matching must be computed in a single pass over already-encoded, evidence-unaware options.

\paragraph{Why interventions work.}
The same formalism predicts which interventions can lift the bottleneck and which cannot:
\begin{itemize}[leftmargin=*,topsep=0pt,parsep=0pt,itemsep=2pt]
    \item \textbf{Option repetition (QOCO).} Repeated option tokens $O'$ placed \emph{after} the context satisfy $h_{O'} = f_\theta(Q, O, C, O')$, restoring $I(h_{O'}; C \mid Q, O) > 0$ and recovering evidence--option alignment. Our QOCO experiment yields $+8.2$ points, the largest zero-intervention-cost gain among interventions.
    \item \textbf{Activation patching.} Replacing $h_O^{\mathrm{QOC}}$ with $h_O^{\mathrm{CQO}}$ directly injects context-conditioned option states, yielding $+6.0$ points.
    \item \textbf{Chain-of-thought prompting.} CoT appends generated tokens $y_{1:M}$ after the full prompt; each $y_m = g_\theta(h_{Q,O,C,y_{<m}})$ is context-aware and effectively performs additional evidence--option comparisons beyond the single final step, reducing the gap from $14.72$ to $7.47$ ($-7.25$ points).
    \item \textbf{Bidirectional architectures.} Encoder-only and encoder--decoder models drop the constraint~\eqref{eq:causal_constraint} and therefore admit $h_O = f_\theta(Q, O, C)$ under both templates, which is why their CQO--QOC gap is essentially zero (\Cref{fig:fig5-a}).
\end{itemize}

\paragraph{Remark.}
Equation~\eqref{eq:mi_qoc} is a structural constraint on the \emph{representation}, not on the final prediction: an oracle single-step matcher $g_\theta$ could in principle compensate for context-blind options. Our experiments show that current decoder-only LMs do not achieve this, and the residual CoT gap ($+7.47$) quantifies how much evidence--option alignment can still be recovered at inference time with additional decoding steps but without changing the causal mask itself.

\section{Generalization beyond MCQA: open-domain QA and RAG}
\label{app:beyond_mcqa}

The main paper focuses on MCQA with four options. To test whether the CQO--QOC ordering effect generalizes to open-ended formats, we additionally evaluate two generative tasks in which there are no pre-specified options: (i) open-domain extractive QA on SQuAD~2.0, and (ii) retrieval-augmented generation (RAG) with BM25-retrieved distractor passages. We use the same 10 instruction-tuned decoder-only models for both settings (a subset of the 21 models in the main paper) and evaluate with token-level F1.

\paragraph{Task 1: Open-domain QA (SQuAD 2.0).}
We use the full SQuAD~2.0 test set ($5{,}928$ questions) and compare two template orderings:
(i) \textsc{CQ} (Context $\rightarrow$ Question), the paper's CQO analogue;
(ii) \textsc{QC} (Question $\rightarrow$ Context), the QOC analogue;
and (iii) \textsc{Q} (Question only, no context), as a floor.
Models generate the answer freely and F1 is computed against gold spans; per-model results are in~\Cref{tab:openqa_results}.

\begin{table}[h]
\centering
\small
\setlength{\tabcolsep}{3pt}
\resizebox{\linewidth}{!}{%
\begin{tabular}{lccc|c}
\toprule
\textbf{Model} & \textbf{CQ F1} & \textbf{QC F1} & \textbf{Q F1} & \textbf{Gap} \\
\midrule
Llama-3.1-8B-Instruct & 83.06 & 74.80 & 23.44 & $+8.26$ \\
Llama-3.2-1B-Instruct & 69.04 & 51.34 & 15.18 & $+17.70$ \\
Llama-3.2-3B          & 42.03 & 26.60 & 13.70 & $+15.43$ \\
Llama-3.2-3B-Instruct & 80.33 & 66.97 & 19.47 & $+13.36$ \\
Qwen2.5-0.5B          & 38.84 & 24.90 & 9.84  & $+13.93$ \\
Qwen2.5-0.5B-Instruct & 40.58 & 24.38 & 10.15 & $+16.19$ \\
Qwen2.5-1.5B-Instruct & 49.23 & 48.40 & 15.35 & $+0.84$  \\
Qwen2.5-7B-Instruct   & 61.17 & 46.06 & 20.48 & $+15.11$ \\
gemma-2-2b-it         & 68.12 & 58.38 & 19.71 & $+9.74$  \\
gemma-2-9b-it         & 79.42 & 73.44 & 25.33 & $+5.99$  \\
\midrule
\textbf{Average (10 models)} & \textbf{61.18} & \textbf{49.53} & \textbf{17.26} & $+11.66$ \\
\bottomrule
\end{tabular}
}
\caption{\textbf{Open-domain QA on SQuAD 2.0 also exhibits the CQO-style gap.} Token-level F1 in three prompt orderings. The $+11.66$-point average CQ$-$QC gap shows that the ordering effect is not an artifact of MCQA with fixed options; it persists under free-form generation.}
\label{tab:openqa_results}
\end{table}

\paragraph{Task 2: Retrieval-augmented generation (RAG).}
We simulate a realistic RAG pipeline: for each SQuAD 2.0 question, we retrieve four BM25 distractor passages that are lexically similar to the question but do not contain the answer, and bundle them with the gold passage. The key comparison is where the question sits relative to the passages:
(i) \textsc{GoldFirst}: $[\text{Gold}, \text{Dist}_1, \dots, \text{Dist}_4] \to Q \to A$, with the gold evidence at the start;
(ii) \textsc{QuestionFirst}: $Q \to [\text{Gold}, \text{Dist}_1, \dots, \text{Dist}_4] \to A$, with the question placed before any passage. Per-model results are in~\Cref{tab:rag_results}.

\begin{table}[h]
\centering
\small
\setlength{\tabcolsep}{3pt}
\resizebox{\linewidth}{!}{%
\begin{tabular}{lcc|c}
\toprule
\textbf{Model} & \textbf{GoldFirst F1} & \textbf{QuestionFirst F1} & \textbf{Gap} \\
\midrule
Llama-3.1-8B-Instruct & 46.48 & 31.16 & $+15.32$ \\
Llama-3.2-1B-Instruct & 39.79 & 19.18 & $+20.62$ \\
Llama-3.2-3B          & 45.38 & 21.34 & $+24.04$ \\
Llama-3.2-3B-Instruct & 31.74 & 21.16 & $+10.58$ \\
Qwen2.5-0.5B          & 29.25 & 20.14 & $+9.12$  \\
Qwen2.5-0.5B-Instruct & 30.45 & 18.51 & $+11.94$ \\
Qwen2.5-1.5B-Instruct & 51.48 & 18.03 & $+33.44$ \\
Qwen2.5-7B-Instruct   & 23.21 & 20.07 & $+3.15$  \\
gemma-2-2b-it         & 43.34 & 27.92 & $+15.42$ \\
gemma-2-9b-it         & 57.67 & 49.85 & $+7.81$  \\
\midrule
\textbf{Average (10 models)} & \textbf{39.88} & \textbf{24.74} & $+15.14$ \\
\bottomrule
\end{tabular}
}
\caption{\textbf{RAG pipelines are even more sensitive to passage placement ($+15.14$ F1 points on average).} \textsc{GoldFirst} places the retrieved passages (gold $+$ BM25 distractors) \emph{before} the question; \textsc{QuestionFirst} places the question first. With four distractors in the bundle, the ordering effect is amplified relative to MCQA (main paper: $+14.67$), suggesting a practical implication: document ordering is a non-trivial design choice for RAG systems.}
\label{tab:rag_results}
\end{table}

\paragraph{Takeaways.}
Both generative tasks exhibit the same ordering effect: placing the evidence \emph{before} the question outperforms the reverse order, with gaps comparable to or larger than those we observe in MCQA. The RAG result ($+15.14$ F1) is particularly striking because distractor passages amplify the bottleneck---the model's question representation is computed without any knowledge of the retrieved passages, and there are five competing passages to compare against at the final answer step. This is consistent with the formalism in Appendix~\ref{app:theory}, where the question (rather than the options) plays the role of the representation that must carry context-aware information through the residual stream.

\end{document}